%% file: draft.tex
\pdfoutput=1

\documentclass[10pt,twocolumn,letterpaper]{article}

\usepackage[margin=2cm]{geometry}

\usepackage{graphicx,epstopdf}
\usepackage{amsmath,eucal,amssymb}
\usepackage{amssymb,bibspacing,url,xspace}

\usepackage[bf,footnotesize]{caption}

\usepackage{mdwtab}
\usepackage{multirow}

\usepackage[scriptsize]{subfigure}

\usepackage[nospace,compress]{cite}
\usepackage{algorithm2e}
\let\savedalgorithm\algorithm
\let\savedendalgorithm\endalgorithm

\usepackage{algorithm}
\newcommand{\fnorm}[2][2]{\ensuremath{ \left\| #2 \right\|^2_{ \mathrm{#1} } } }

\newcommand{\iid}{{i.i.d.}\xspace}
\newcommand{\argmax}{\operatornamewithlimits{argmax}}
\newcommand{\argmin}{\operatornamewithlimits{argmin}}

\newcommand{\half}{{\frac{1}{2}}}

\newcommand{\crf}{{CRFs}\xspace}
\newcommand{\mrf}{{MRFs}\xspace}
\newcommand{\cnn}{{CNNs}\xspace}
\newcommand{\rnn}{{RNNs}\xspace}

\makeatletter
\DeclareRobustCommand\onedot{\futurelet\@let@token\@onedot}
\def\@onedot{\ifx\@let@token.\else.\null\fi\xspace}

\def\eg{\emph{e.g}\onedot} 
\def\ie{\emph{i.e}\onedot} 
 
\def\etc{\emph{etc}\onedot} 
 
\def\etal{\emph{et al}\onedot}
\makeatother

\def\Real{\mathbb{R}}

\def\bregress{{\bf z}}
\def\pws{{R}}
\def\bpws{{\bf R}}

\def\x{{\bf x}}
\def\y{{\bf y}}

\def\I{{\bf I}}

\def\D{{\bf D}}
\def\A{{\bf A}}

\def\loc{{\bf l}}
\def\pout{{\bf s}}
\def\btheta{{\boldsymbol \theta}}

\def\butheta{{\boldsymbol \theta^U}}
\def\bptheta{{\boldsymbol \theta^V}}

\def\T{{\mathrm{T}}}
\def\Pr{{\mathrm{Pr}}}

\def\calN{{\cal N}}
\def\calS{{\cal S}}
\def\calL{{\cal L}}
\def\calD{{\cal D}}

\def\T{{\!\top}}

\begin{document}

\title{Discriminative Training of Deep Fully-connected Continuous \crf \\
with Task-specific Loss}

\author{Fayao Liu, Guosheng Lin, Chunhua Shen\thanks{Corresponding author (e-mail: chhshen@gmail.com). }
\\
School of Computer Science,
University of Adelaide, Australia}

\maketitle
\thispagestyle{empty}

\input{intro.tex}

\input{related_work.tex}

\input{method.tex}

\begin{figure*} \center
\resizebox{.7\linewidth}{!} {
\begin{tabular}{cccc}

      \includegraphics[width=0.3\textwidth, height=0.22\textwidth]{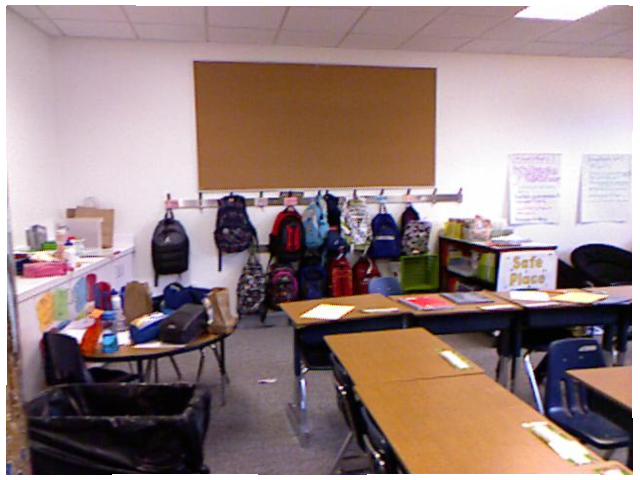}
      &\includegraphics[width=0.3\textwidth, height=0.22\textwidth]{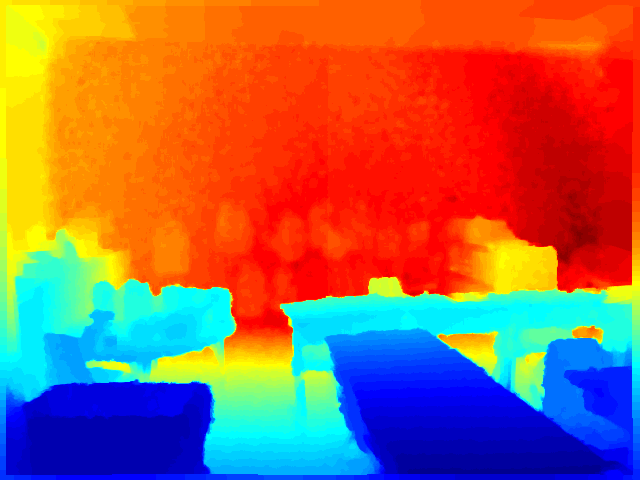}
      &\includegraphics[width=0.3\textwidth, height=0.22\textwidth]{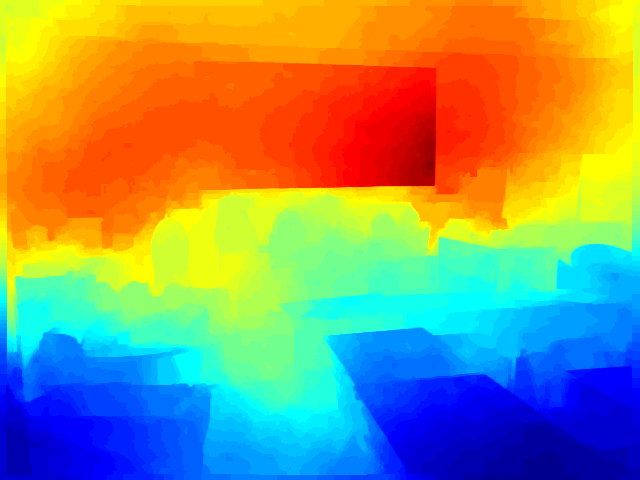}   \\

      \includegraphics[width=0.3\textwidth, height=0.22\textwidth]{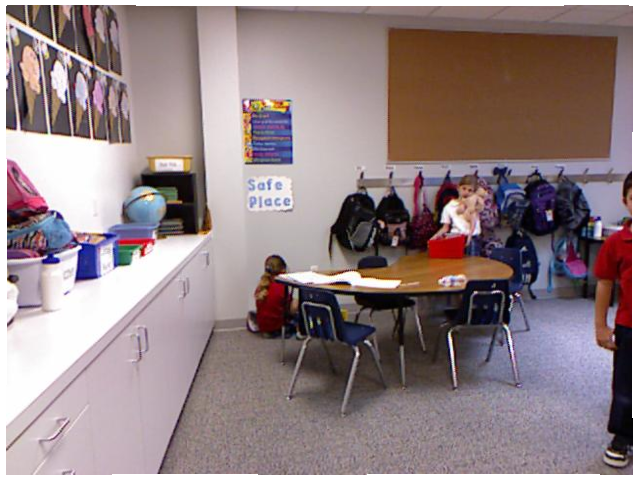}
      &\includegraphics[width=0.3\textwidth, height=0.22\textwidth]{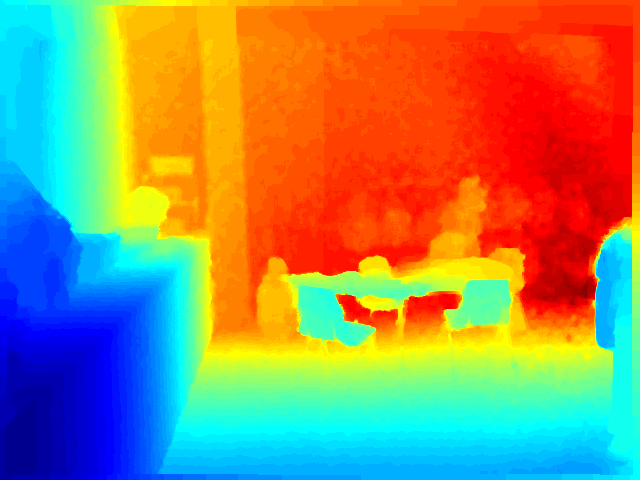}
      &\includegraphics[width=0.3\textwidth, height=0.22\textwidth]{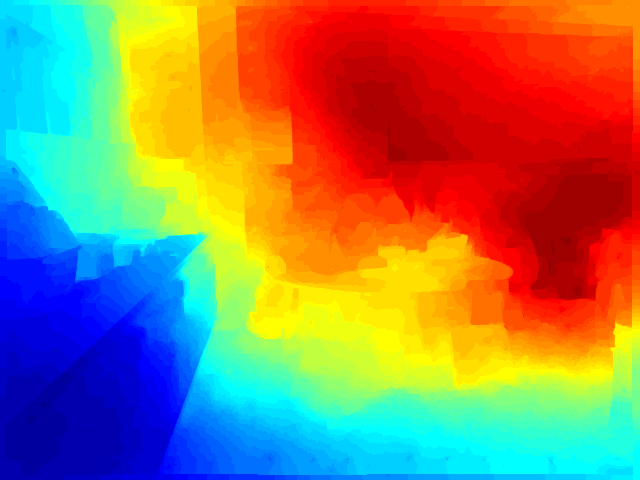}   \\

      \includegraphics[width=0.3\textwidth, height=0.22\textwidth]{./fig/examples/nyud/img/0330.png}
      &\includegraphics[width=0.3\textwidth, height=0.22\textwidth]{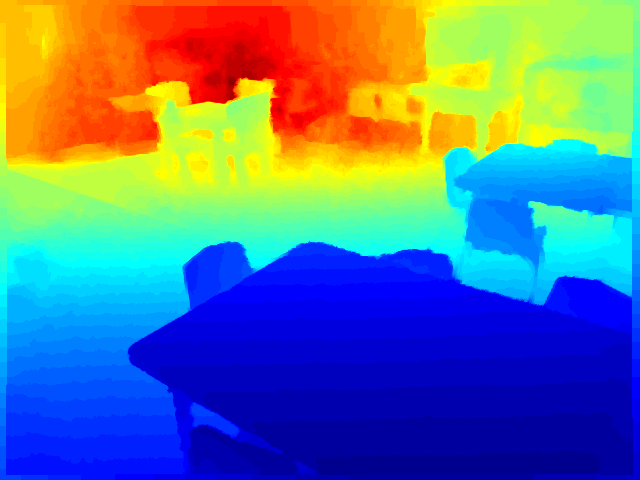}
      &\includegraphics[width=0.3\textwidth, height=0.22\textwidth]{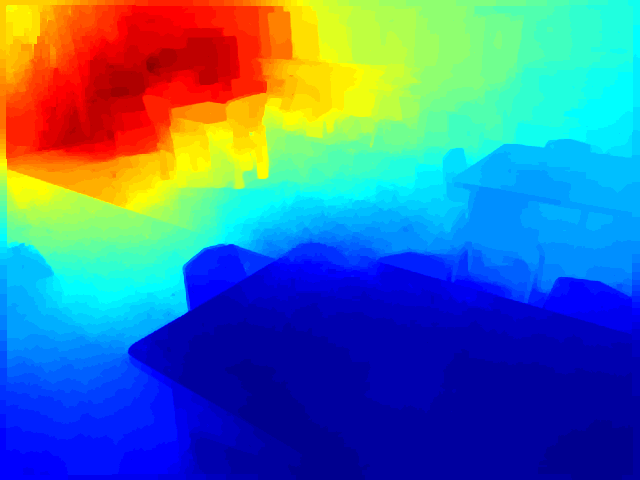}   \\

      \includegraphics[width=0.3\textwidth, height=0.22\textwidth]{./fig/examples/nyud/img/0976.png}
      &\includegraphics[width=0.3\textwidth, height=0.22\textwidth]{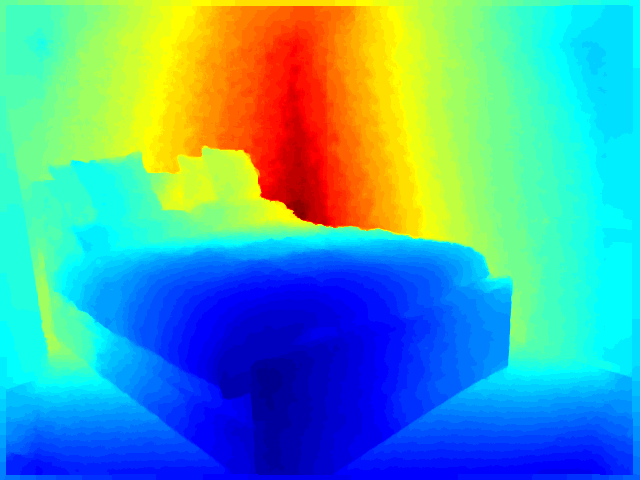}
      &\includegraphics[width=0.3\textwidth, height=0.22\textwidth]{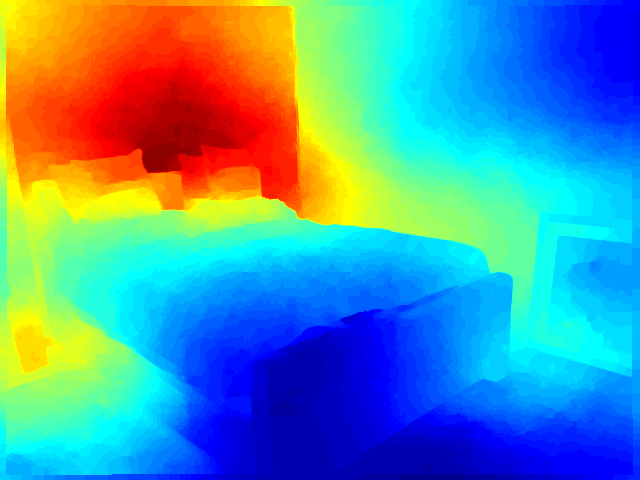}   \\

	\includegraphics[width=0.3\textwidth, height=0.22\textwidth]{./fig/examples/nyud/img/1082.png}
      &\includegraphics[width=0.3\textwidth, height=0.22\textwidth]{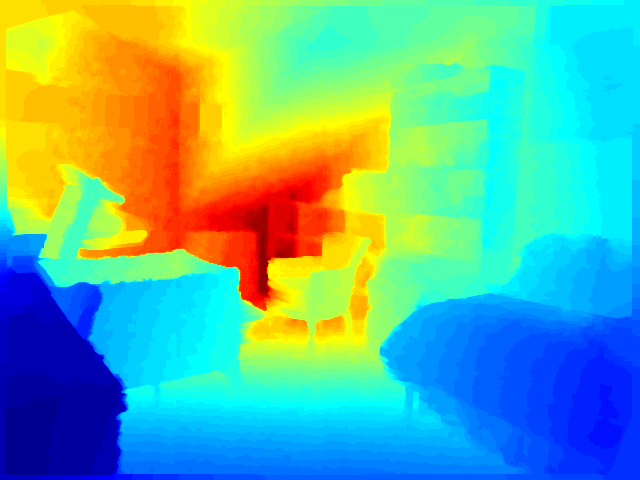}
      &\includegraphics[width=0.3\textwidth, height=0.22\textwidth]{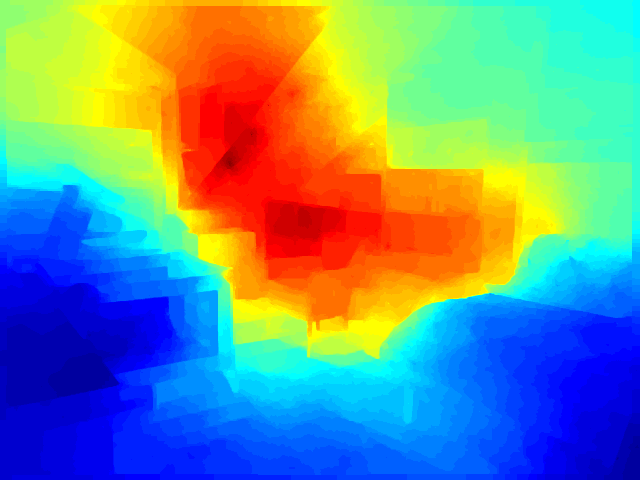}   \\

	\includegraphics[width=0.3\textwidth, height=0.22\textwidth]{./fig/examples/nyud/img/1148.png}
      &\includegraphics[width=0.3\textwidth, height=0.22\textwidth]{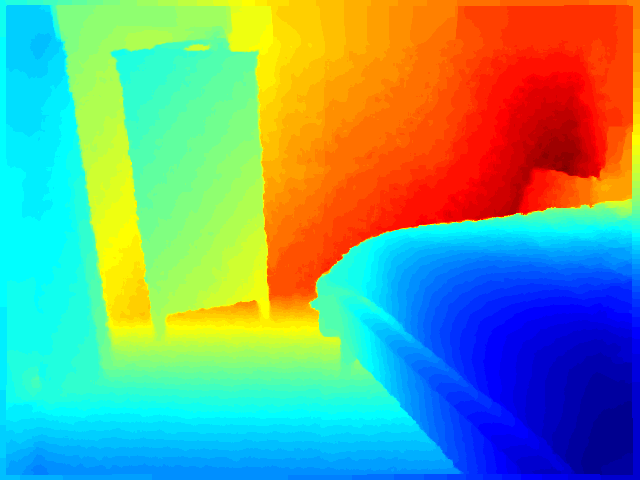}
      &\includegraphics[width=0.3\textwidth, height=0.22\textwidth]{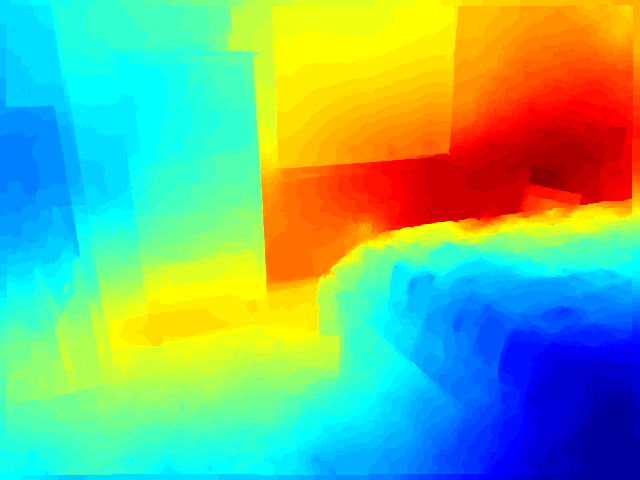}   \\

      \includegraphics[width=0.3\textwidth, height=0.22\textwidth]{./fig/examples/nyud/img/1292.png}
      &\includegraphics[width=0.3\textwidth, height=0.22\textwidth]{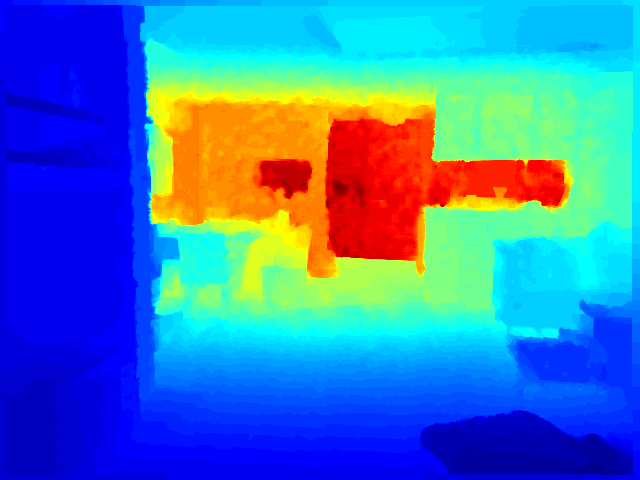}
      &\includegraphics[width=0.3\textwidth, height=0.22\textwidth]{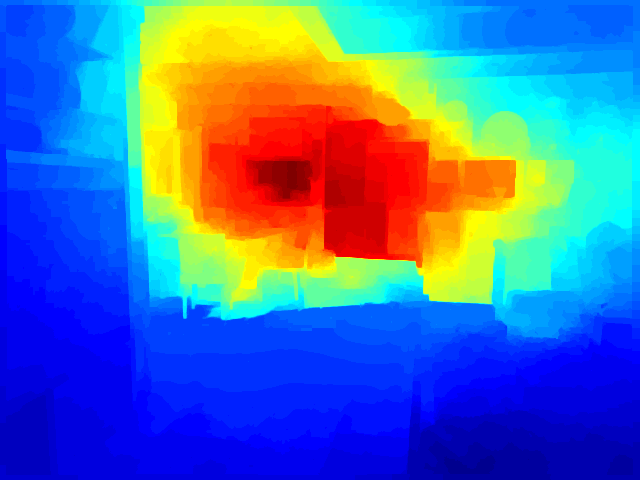}   \\

      \includegraphics[width=0.3\textwidth, height=0.22\textwidth]{./fig/examples/nyud/img/1314.png}
      &\includegraphics[width=0.3\textwidth, height=0.22\textwidth]{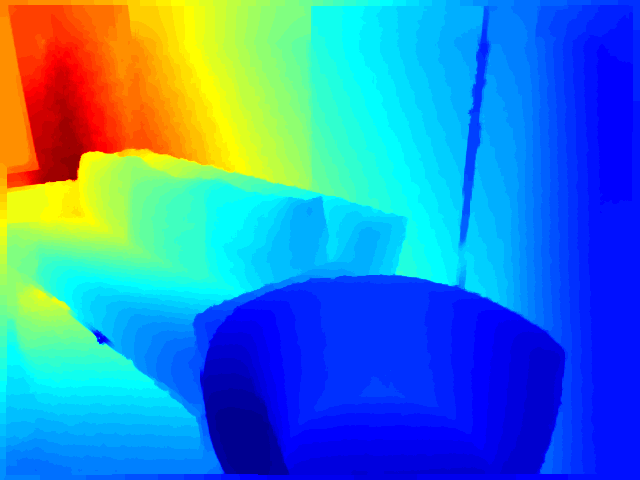}
      &\includegraphics[width=0.3\textwidth, height=0.22\textwidth]{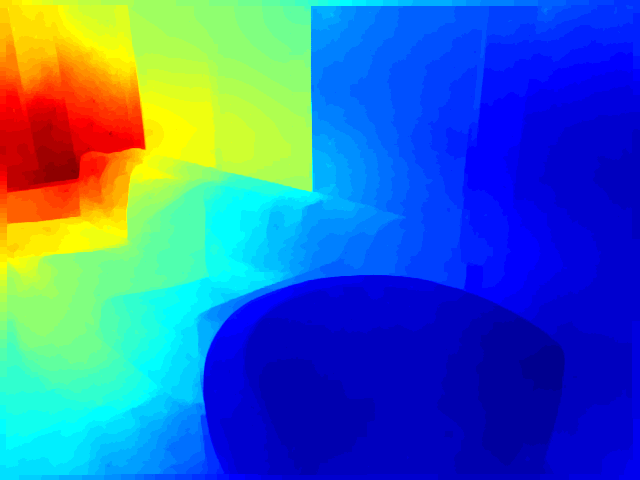}   \\

Test image &Ground-truth  &Our prediction \\

\end{tabular}
}
\caption{Depth prediction examples of our method on the NYU v2 dataset. }  \label{fig:nyud2_suppl}
\end{figure*}

\input{exp.tex}
We show some depth prediction examples on the NYU v2 dataset in Fig. \ref{fig:nyud2_suppl}.
As we can see, our predictions well preserve depth discontinuities and align to local details.

\section{Conclusion}
We have proposed a fully-connected deep continuous \crf learning method for both discrete and continuous pixel-level labelling problems.
The proposed method inherits the advantages of continuous \crf, \ie, exact maximum likelihood learning and closed-form solution for the MAP inference.
Specifically, it models the unary and the pairwise potentials of the \crf as two \cnn and learns the network parameters in an end-to-end fashion.
For different tasks,  we further propose to optimize a task-specific loss function, rather than the traditional log-likelihood loss in learning \crf parameters,  and demonstrate two applications, which are semantic labelling and robust depth estimation.
Experimental results show that (a) the proposed method improves the unary-only model (standard regression without considering structured constraints), indicating the effectiveness of the pairwise term of our model;
(b) our method achieves better or on par results compared with state-of-the-art, even on discrete labelling tasks,
showing the promise of the proposed deep structured model.
    We thus advocate the use deep continuous \crf in more structured prediction problems in vision, due
    to its power and in particular, significantly simpler and efficient inference, compared with most discrete deep \crf such as \cite{Lin15a,Lin15b}.

{\small
   \bibliographystyle{ieee-cs}
   \bibliography{seg.bib}
}

\end{document}

%% file: intro.tex
\begin{abstract}
Recent works on deep conditional random fields (\crf) have set new records on
many vision tasks involving structured predictions.
Here we propose a fully-connected deep continuous \crf model for both discrete and continuous labelling
problems. We exemplify the usefulness of the proposed model on
multi-class semantic labelling (discrete) and the robust depth estimation (continuous) problems.
 In our framework, we model both the unary and the pairwise potential functions as deep convolutional neural networks (\cnn), which are jointly learned in an end-to-end fashion.
The proposed method possesses the main advantage of continuously-valued \crf, which is a closed-form solution for the Maximum a posteriori (MAP) inference.

 To better adapt to different tasks, instead of using the commonly employed maximum likelihood \crf parameter learning protocol,
 we propose task-specific loss functions for learning the \crf parameters.
 It enables direct optimization of the quality of the MAP estimates during the course of learning.
 Specifically, we optimize the multi-class classification loss for the semantic labelling task and the Turkey's biweight loss for the robust depth estimation problem.
 Experimental results on the semantic labelling and robust depth estimation tasks demonstrate that the proposed method compare favorably against both baseline and state-of-the-art methods.
 In particular, we show that although the proposed deep \crf model is continuously valued,
 with the equipment of task-specific loss, it achieves impressive results even on discrete labelling tasks.
\end{abstract}

\section{Introduction}

Recent works on combining conditional random fields (\crf) and deep convolutional neural networks (\cnn)
have set new records on many vision tasks involving structured predictions.
Pixel-level labelling tasks generally refer to assigning a discrete or continuous label to each pixel in an image, with typical examples being semantic labelling, and depth estimation.
They play a pivotal role in the computer vision community, and are fundamental to many high-level tasks, like recognition, scene understanding, 3D modeling.
In the central of many pixel labelling tasks stands the feature engineering, which has a significant impact on the final results.
However, designing effective and highly discriminative features remains a challenge.
Traditional efforts have been focusing on designing hand-crafted features, \eg, texton, HOG, SIFT, GIST, \etc.
Later on, unsupervised feature learning \cite{Coates11} has been introduced and showed promises.
Recently, supervised feature learning, \ie, deep  \cnn have achieved  tremendous success in various domains including image classification \cite{AlexNet12}, pose estimation \cite{Lecun_nips14}, semantic labelling \cite{Long15,Zheng15,Eigen15}, depth estimation \cite{dcnn_nips14,cvpr15}, \etc.
Adapting \cnn for pixel labelling tasks have since attracted a lot of attention.
However, several issues need to be resolved in this procedure, \eg, down-sampled coarse predictions, non-sharp boundary transitions.

Besides the feature representation, another essential factor to be considered is the appearance and spatial consistency, which is typically captured by graphical models.
Probabilistic graphical models, \eg, \crf \cite{Lafferty_icml01}, have long been shown as a fundamental tool in computer vision, and have been widely applied to structured prediction problems, like image denoising, semantic labelling, depth estimation, \etc.
Recently, continuous \crf  have shown benefits of efficient learning and inference, while being as powerful as discrete \crf models \cite{Tappen08,Samuel09,CCNF_eccv14,cvpr15,pami15}.
In this paper, we present a general deep \crf learning framework for both discrete and continuous labelling problems, which combines the benefits of \cnn and continuous \crf.

While the maximum likelihood learning is dominant in learning \crf parameters, it is argued being beneficial to directly take into account the quality of the MAP estimates during the course of learning.
 Samuel  \etal \cite{Samuel09} pointed out that if one's goal is to use MAP estimates and evaluate the results using other criteria, then optimization of the \mrf parameters using a probabilistic criterion, like maximum likelihood, may not necessarily lead to optimal results.
Instead, a better strategy is to find the parameters such that the quality of the MAP estimates are directly optimized against a task-specific empirical risk.
For example, in  image segmentation, the most direct criterion of a good segmentation mask is the global accuracy or the Jaccard index (intersection-over-union score).
On the other hand, in many practical applications, it may  not be necessary to model the full conditional distribution of labels, given observations.
Instead, modeling the marginal conditional likelihood is more preferable.
In \cite{Domke13}, Domke argued that the marginal based loss generally yields better
maximum posterior marginal
(MPM) inference.
Therefore an alternative to the maximum likelihood learning is to minimize the  empirical risk loss.
In \cite{Kakade02}, Kakade \etal propose to maximize the label-wise marginal likelihood
rather than the joint label likelihood for sequence labelling problems.
Here we analogously propose to maximize the label-wise labelling accuracy by optimizing the softmax classification loss for the semantic labelling task.
For the depth estimation problem, depths captured by  depth sensors are often noisy and contain missing values due to poor illumination and sensor limitation, \etc.
We  tackle this problem by proposing a robust depth estimation model, which uses the robust Turkey's biweight loss.

We summarize the main contributions of this work as follows.
\begin{itemize}
\itemsep -.12cm
\item We present a general deep continuous \crf learning model for both discrete and continuous pixel-level labelling problems.
Both the unary and the pairwise potentials are modeled as \cnn networks, and jointly learned in an end-to-end fashion. There exists a closed-form solution for the MAP inference problem, largely reducing the complexity of the model.
Our \crf model is fully connected and thus may better capture long-range relations.
\item Instead of the commonly used maximum likelihood criterion for \crf parameter learning, we show that a task-specific loss function can be applied to directly optimize the MAP estimates, and show two different applications, namely, multi-class semantic labelling (discrete) and robust depth estimation (continuous).
\item We experimentally demonstrate that the proposed framework shows advantages over the baseline methods and state-of-the-art methods on both semantic labelling and robust depth estimation tasks.
\end{itemize}

There seems to be a common understanding that for discrete labelling problems,  continuous \crf models  may not be suitable.  As a result, an increasingly sophisticated and computationally expensive  discrete  \crf methods, in terms of both training and inference, have been developed in the literature. Here we argue that {\em a carefully-designed, yet much simpler, continuous \crf model is capable of solving discrete labelling problems effectively and producing competitive results, at least in the case of  deep  \crf}---\crf with deep \cnn modeling the potentials.

%% file: related_work.tex
\subsection{Related work}

In this section, we review some most recent progresses in \cnn based methods for semantic labelling and depth estimation tasks, which are closely related to our work.

\paragraph{\cnn for semantic labelling}
The most straightforward way of applying \cnn for semantic labelling is to design fully convolutional networks to directly output the prediction map, \eg, \cite{Long15,Eigen15}.
These methods do not involve optimized structured loss, while only consider independent classification.
A more appealing direction is to combine \cnn with graphical models, which explicitly incorporates structured constraints.
A simple strategy to do so is to incorporate consistency constraints into a trained unary model as a post-processing step. These methods first train a \cnn model or generate \cnn features for constructing the unary potential, then add spatial pairwise constraints to optimize the \crf loss.
In the pioneering work of \cite{Farabet13}, Farabet \etal present a multi-scale \cnn framework for scene labelling, which uses \crf as a post-processing step for local refinement.
Most recently, Chen \etal \cite{deeplab} propose to first train a fully convolutional neural networks for pixel classification and then separately apply a dense \crf to refine the semantic labelling results.
Later on, more attempts are made towards joint learning of \cnn and graphical models.
In the work of \cite{Zheng15}, Zheng \etal propose to implement the mean field inference in \crf as recurrent neural networks (\rnn) to facilitate the end-to-end joint learning.
Lin \etal \cite{Lin15a} propose a piecewise training approach to learn the unary and pairwise \cnn potentials in \crf for semantic labelling.
They further propose in \cite{Lin15b} to directly learn \cnn message estimators in the message passing inference rather than learning the potential functions.
All of the above-mentioned joint learning approaches exploit discrete \crf, which need to design approximation methods for both learning and inference.
In contrast, we here explore continuous \crf, which enjoys the benefits of exact maximum likelihood learning and closed-form solution for the MAP inference.

\paragraph{\cnn for depth estimation}
For the depth estimation task, Eigen \etal \cite{dcnn_nips14} propose to
train multi-scale \cnn to directly output the predicted depth maps by optimizing the pixel-wise least square loss for depth regression.
During the course of network training, there is no explicit structured constraints involved.
Most recently, Liu \etal \cite{cvpr15} propose a deep convolutional neural fields model for depth estimation from single images, which jointly learns continuous \crf and deep \cnn in a single framework.
They further propose a more effcient training approach based on fully convolutional networks
and a superpixel pooling method in \cite{pami15}.
Our work here, which also jointly learn continuous \crf and \cnn in a single framework,
is mainly inspired by these two works \cite{cvpr15,pami15}.
However, our work differs from \cite{cvpr15,pami15} in the following important aspects.
(a) Our method  addresses both discrete and continuous labelling problems, while the works in \cite{cvpr15,pami15} only deal with continuous labelling problem;
Apart from continuous prediction, we extend continuous deep \crf with closed-form MAP inference for discrete prediction.  We show that by optimizing the task-specific empirical risk, it is possible to achieve competitive  performance for discrete labelling tasks.
(b) In contrast to the maximum likelihood optimization in \cite{cvpr15,pami15},
we show the flexibility and usefulness of employing a task-specific loss in the context of deep \crf.
(c) Our method here is based on fully-connected \crf with both the unary and the  pairwise potentials modeled as deep CNNs. In contrast,  the pairwise term  in \cite{cvpr15,pami15} is a linear function  (or single layer) built upon
hand-crafted image features. In our model,  the pairwise term is also a \cnn model, and all the features are learned end-to-end.

%% file: method.tex
\section{Fully-connected deep continuous CRFs}
\begin{figure*}  \center
      \includegraphics[width=0.79\textwidth]{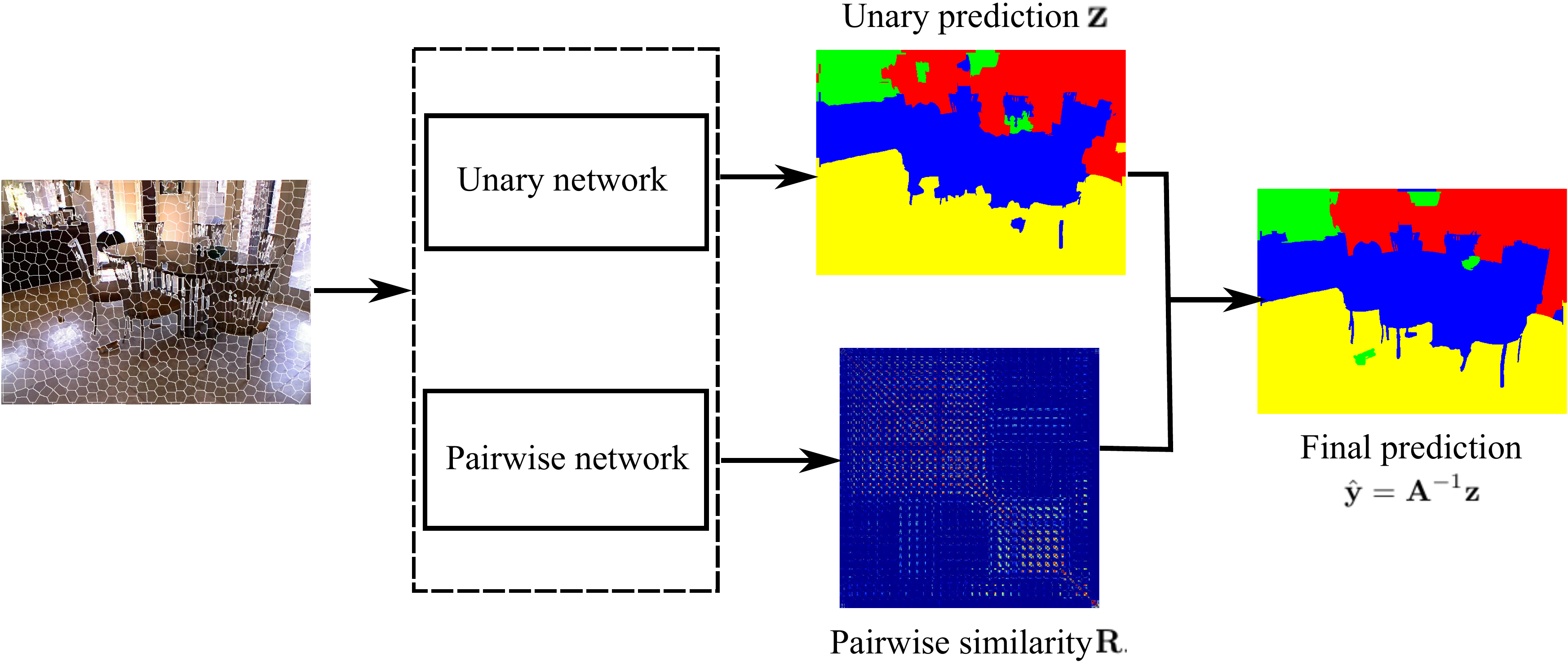}
\caption{The prediction framework of our method for the semantic labelling task. The image is first over-segmented into superpixels, and then as input to the unary and the pairwise networks. The unary network outputs the unary prediction of each superpixel, and the pairwise network outputs the pairwise similarities of superpixe pairs. The final prediction is obtained by combining the outputs of the unary and the pairwise networks to solve the MAP inference (Eq. \eqref{eq:inf_map}). Details of the unary and the pairwise networks are illustrated in Fig. \ref{fig:cnn_net}. } \label{fig:dcnf_seg}
\end{figure*}

Fig.~\ref{fig:dcnf_seg} shows an overview of our
fully-connected deep continuous \crf model, which consists of a unary network and a pairwise network.
The unary network generates the unary prediction and is parametrized by $\butheta$,
while the pairwise network outputs a similarity matrix of superpixel pairs and is parametrized by $\bptheta$.
The final prediction is performed by combining the unary and the pairwise outputs to conduct the MAP inference.

\subsection{Discriminative learning of model parameters}
Given $N$ \iid training examples $\calD=\{(\x^{(i)}, \y^{(i)})\}_{i=1}^N$, \crf typically learn the model parameters $\btheta$ by the maximum likelihood learning, \ie, minimizing the negative conditional log-likelihood:
\begin{align} \label{eq:opt_crf}
\min_{\btheta} \frac{\lambda}{2} \fnorm \btheta - \sum_{i=1}^N \log \Pr(\y^{(i)}|\x^{(i)}; \btheta),
\end{align}
where $\frac{\lambda}{2} \fnorm \btheta$ is the regularization term, with $\lambda$ being the weight decay parameter.
The conditional log-likelihood is modelled as:
\begin{equation}\label{eq:prob}
\begin{aligned}
\Pr(\y|\x) = \frac{1}{\mathrm{Z(\x)}} \exp (- E(\y, \x)),
\end{aligned}
\end{equation}
where $E$ is an energy function as described in Sec.~\ref{sec:energy}; $\mathrm{Z}$ is the partition function defined as:
$\mathrm{Z}(\x) = \int_{\y} \exp \{ -E(\y, \x) \}\mathrm{d}\y$ for continuous labelling $\y$.
During prediction, one performs the MAP inference:
\begin{equation}\label{eq:inf_map_org}
\begin{aligned}
\hat{\y} = \argmax_{\y} \Pr(\y|\x; \btheta).
\end{aligned}
\end{equation}

Instead of the traditionally optimized log-likelihood objective, we propose to directly optimize the quality of the estimated labelling (MAP solution).
Similar to the discriminative max-margin framework proposed by Taskar \etal \cite{Taskar05}, our learning objective is:
\begin{align} \label{eq:opt_all}
\min_{\btheta} &\frac{\lambda}{2} \fnorm \btheta + \sum_{i=1}^N \calL(\hat{\y}^{(i)}, \y^{(i)}),  \notag \\
\text{where}\;\; & \hat{\y}^{(i)} = \argmax_{\y} \Pr(\y|\x^{(i)}; \btheta).
\end{align}
Here $\calL(\cdot, \cdot)$ is a task-specific loss function, which measures the discrepancy of the predicted label with the ground-truth label;  $\y^{(i)}$, $\hat{\y}^{(i)}$ are respectively the ground-truth and the predicted labels of the $i$-th training example.
The loss function $\calL$ provides important information on how good a potential prediction $\hat{\y}^{(i)}$ is with respect to the ground-truth label $\y^{(i)}$.
This is generally not addressed in the traditional \crf learning approaches that rely on the maximum likelihood scheme.
To our knowledge, this has not been explored in the case of learning {\em deep} CRFs parameters.
Note that
the implicit differentiation technique can be applied to
compute the gradient of  $  \calL(\hat{\y}^{(i)}, \y^{(i)}) $     with respect to the parameters
$  \btheta $, which enables us to use stochastic gradient descent (SGD) to learn deep structured models.

\begin{figure*}[t!]  \center
      \includegraphics[width=0.76\textwidth]{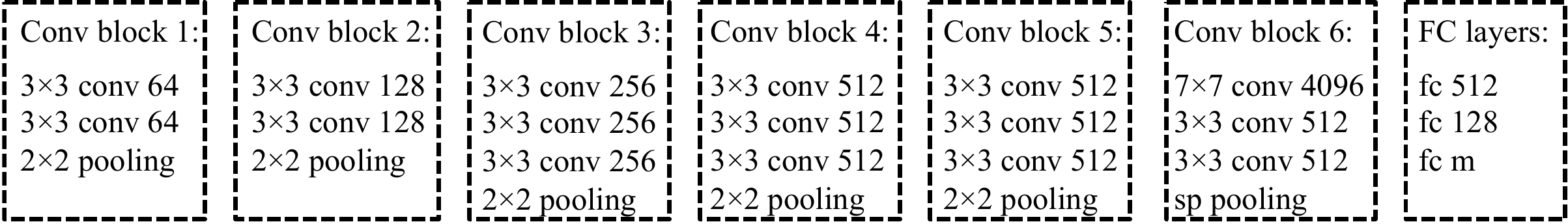} \\
      \includegraphics[width=0.45\textwidth]{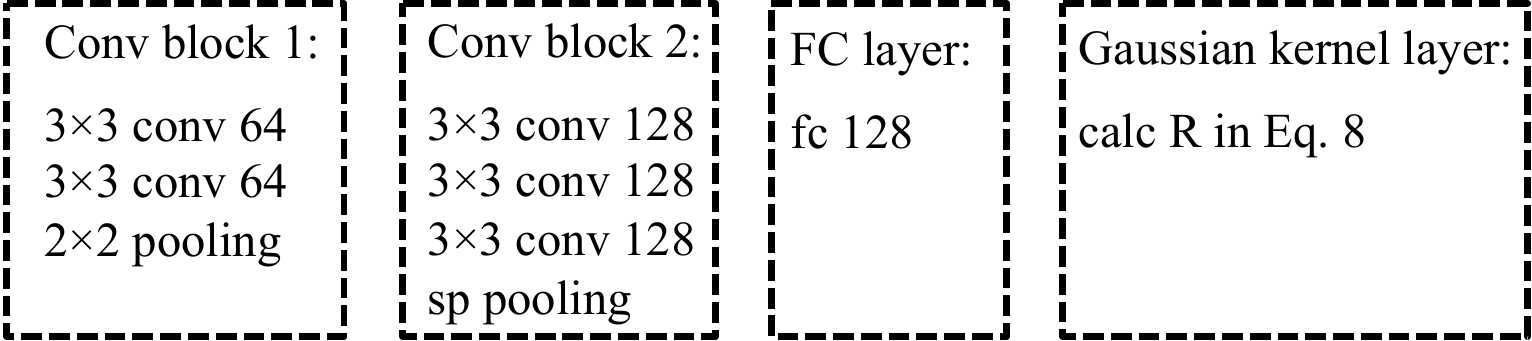}
\caption{The unary (upper plot) and the pairwise (lower plot) network architectures of our method.
The unary network is composed of $6$ convolution blocks and $3$ fully-connected layers.
The pairwise network is relatively shallow, which consists of $2$ convolution blocks together with one fully-connected layer and a Gaussian kernel layer to calculate the pairwise similarity matrix $\bpws$ in Eq. \eqref{eq:def_R}.
$m$ is the number of semantic classes. We use the superpixel pooling (denoted as sp pooling layer) method as proposed in \cite{pami15} to obtain superpixel \cnn features from the convolution maps.}  \label{fig:cnn_net}
\end{figure*}

\subsection{Energy formulation}
\label{sec:energy}
We now show how to generalize the deep continuous \crf model in \cite{pami15} to fully-connected multi-label discrete labelling problems, \eg, multi-class image segmentation.

For the multi-class image segmentation task, given an image $\x$, we represent the label of each node (superpixel) as an $m$ dimensional vector $\y_p$ ($\y_p \in \Real^m$), which consists of the confidence scores of superpixel $p$ being each of the $m$ classes. Here $m$ is the total number of semantic classes.
Let $\y=\y_1 \odot \y_2 \odot \ldots \odot \y_n$ denotes the overall segmentation label of the image $\x$, with $\odot$ stacking two vectors and $n$ being the number of superpixels in $\x$.
We formulate the energy function as a typical combination of unary potentials and pairwise potentials over the nodes (superpixels) $\calN$ and edges $\calS$ of the image $\x$:
\begin{equation}\label{eq:energy}
E(\y, \x) = \sum_{p \in \calN} U(\y_{p}, \x) + \sum_{(p,q) \in \calS} V(\y_{p}, \y_{q}, \x),
\end{equation}
where $U$, $V$ are the unary and pairwise potential function respectively.

The unary potential is formulated as:
\begin{equation}\label{eq:unary}
U(\y_{p}, \x;\butheta) = \fnorm{\y_p - \bregress_p}. %
\end{equation}
Here $\bregress_p$ is the unary network (network architecture detailed in Fig. \ref{fig:cnn_net}) output for the superpixel $p$, which is parametrized by $\butheta$; $\fnorm{\cdot}$ denotes the squared $\ell_2$ norm of a vector.

The pairwise potential is written as:
\begin{align} \label{eq:pairwise}
&V(\y_{p}, \y_{q}, \x; \bptheta) = \half \pws_{pq} \fnorm{\y_p - \y_q}. %
\end{align}
Here $\pws_{pq}$, parameterized by the pairwise network parameters $\bptheta$,  is the output of the pairwise network (network architecture detailed in Fig. \ref{fig:cnn_net}) from the superpixel pair $(p,q)$. Specifically, $\pws_{pq}$ is computed from a Gaussian kernel layer:\footnote{Note that we are not limited to the Gaussian kernel here. We can use other similarity calculation forms such as Laplacian.
}
\begin{align} \label{eq:def_R}
\pws_{pq} (\bptheta)=\beta \exp ( - \fnorm{\pout_p - \pout_q } - \gamma \fnorm{\loc_p - \loc_q } ),
\end{align}
where $\pout_p$, $\pout_q$  are the convolution features (with dimension being $128$  as illustrated in Fig. \ref{fig:cnn_net}) for superpixel $p$, $q$ respectively; $\loc_p$, $\loc_q$ are the centroid coordinates of the superpixel $p$ and $q$ respectively (normalized to [0, 1]);
$\gamma$ is a hyperparameter that controls the scale of the Gaussian position kernel; $\beta$ is a non-negative scaling factor, which is regarded as one of the  pairwise network parameters and is jointly learned in our framework.

With Eqs.~\eqref{eq:unary} and  \eqref{eq:pairwise}, the energy function in Eq. \eqref{eq:energy} can now be written as:
\begin{align}\label{eq:energy_expand}
E(\y, \x) &= \sum_{p \in {\cal N} } \fnorm{\y_p - \bregress_p}
+ \sum_{(p,q) \in {\cal S}} \half \pws_{pq} \fnorm{\y_p - \y_q}  \notag  \\
&=\y^\T \A \y - 2\bregress^\T \y + \bregress^\T \bregress,
\end{align}
where
\begin{align}
\A &=\A_0 \otimes \I_m, \\
\text{with}\;\; \A_0 &= \I_n + \D - \bpws.
\end{align}
Here $\bregress=\bregress_1 \odot \bregress_2 \odot \ldots \odot \bregress_n$;
$\I_m$, $\I_n$ are identity matrices of size $m \times m$, $n \times n$ respectively;
$\otimes$ denotes the Kronecker product.
$\bpws$ is the $n \times n$ matrix composed of $\pws_{pq}$;
$\D$ is a diagonal matrix with $D_{pp}=\sum_q \pws_{pq}$.

\subsection{Optimization}

\textbf{MAP Inference}
To predict the label of a new given image $\x$, we perform the maximum a posterior (MAP) inference, which is:
\begin{align} \label{eq:inf_map}
\hat{\y}&=\argmax_{\y} \Pr(\y|\x) = \argmin_{\y} E(\y, \x).
\end{align}
With the energy formulation in Eq. \eqref{eq:energy_expand}, we can obtain the closed-form solution for Eq. \eqref{eq:inf_map}:
\begin{align} \label{eq:inf_solution}
\hat{\y}&=\A^{-1} \bregress,
\end{align}
where $\A^{-1}=(\A_0 \otimes \I_m)^{-1} =\A_0^{-1} \otimes \I_m$, with $\A_0^{-1}$ can be obtained by solving a linear equation system. Note that for computational efficiency, one typically does not need to compute
the matrix inverse explicitly.

{\it Discussion on MAP inference of fully connected \crf}
    Note that even for fully connected models, the MAP inference here can still be
    calculated efficiently, due to the closed form of Eq. \eqref{eq:inf_solution}.
    In contrast in the case of discrete fully-connected \crf models,
     efficient inference, even with various approximations,  is only possible for a very special case to date.
      As indicated in
    \cite{krahenbuhl2012efficient}, in order to apply the approximated fast inference of \cite{krahenbuhl2012efficient},
    the pairwise term must be in the form of the Gaussian kernel, and the input features must be in low dimension, typically less than $20$.
    In our case, the input feature's dimension is $128$, which is much higher than what can be possibly handled by
    \cite{krahenbuhl2012efficient}.

\subsubsection{Maximum likelihood learning}
Since here we  consider the continuous domain of $\y$ (vector of confidence scores), the integral in the partition function in
Eq.~\eqref{eq:prob} can be analytically calculated
under certain circumstances, as shown in \cite{pami15}.
According to \cite{pami15}, the negative conditional log-likelihood can be written as:
\begin{align} \label{eq:log-likelihood}
-\log\Pr(\y|\x) =&\y^\T \A \y - 2\bregress^\T \y + \bregress ^\T \A^{-1} \bregress \notag \\
&- \half\log(|\A|) + \frac{n}{2}\log(\pi).
\end{align}
The gradients of  $-\log\Pr(\y|\x)$ with respect to $\A$ and $\bregress$ can be analytically calculated as shown in \cite{pami15}.
Then applying the chain rule (back propagation), the gradients with respect to $\butheta$, $\bptheta$ and $\beta$ can be calculated accordingly.

\subsubsection{Task specific learning }

\paragraph{Multi-class semantic labelling}
In the multi-class semantic labelling task, we obtain
 a prediction of $m$ dimensional vector $\hat{\y}_p$ for superpixel $p$, with each $\hat{y}_{pj}$ ($j$-th element of $\hat{\y}_p$, $j=1,2,\ldots,m$) indicating the confidence score of the superpixel $p$ belonging to the $j$-th category.
 For the training ground truth $\y_p$, the confidence value for the ground truth class is $1$, and $0$ for other classes.
During optimization, we prefer the confidence score of the ground-truth category being as large as possible compared to the confidence scores of incorrect categories.
In this circumstance, we  minimize the  softmax loss, typically employed in multi-class classification,
 \ie, maximize the label-wise classification accuracy, to directly pursue a desirable global accuracy:

\begin{align} \label{eq:opt_seg}
 \calL(\hat{\y}, \y) = - \sum_{p \in \calN} \sum_{j=1}^m y_{pj} \log \frac{\exp(-\hat{y}_{pj})}{\sum_{k=1}^m \exp(-\hat{y}_{pk})}.
\end{align}
Then after we get the prediction $\hat{\y}_p$, the predicted category label for the $p$-th superpixel is:
\begin{align} \label{eq:pred}
 \hat{t}=\argmax_j \hat{y}_{pj}.
\end{align}
In the experiment section,
we show that the use of this classification loss, instead of the traditional maximum likelihood loss,
 is critically important in achieving good performance for multi-class labelling tasks with our deep continuously-valued  \crf model.

\paragraph{Robust depth regression}
For the depth estimation problem, one of the main criteria for a good predictions is the root mean square (rms) error.
Therefore, we can train our model by directly pursuing good MAP estimates in terms of RMS measure.
To achieve this, the loss function for the depth estimation problem can be written as:
\begin{align} \label{eq:opt_depth}
 \calL(\hat{\y}, \y) =\sum_{p \in \calN} \rho(r_p) ,
\end{align}
where $r_p=y_p-\hat{y}_p$  is the depth residual of the superpixel $p$ regarding to the ground truth;
$\rho(\cdot)$ is a penalty function, which generally satisfies $\rho(0)=0$ and $\rho(r_p) > 0$ for $r_p \neq 0$.

A straightforward alternative to the log-likelihood loss is the least squares (LS) loss:
\begin{align} \label{eq:lsloss}
 \rho(r_p)  = r_p^2,
\end{align}
However,   LS loss is generally not robust to outliers. Actually all convex regression loss functions are considered not robust.
For example, for those extremely large residuals (typically outliers), the LS loss imposes too large penalty, which inevitably biases the model towards outliers.

\begin{figure} \center
   \includegraphics[width=0.2\textwidth]{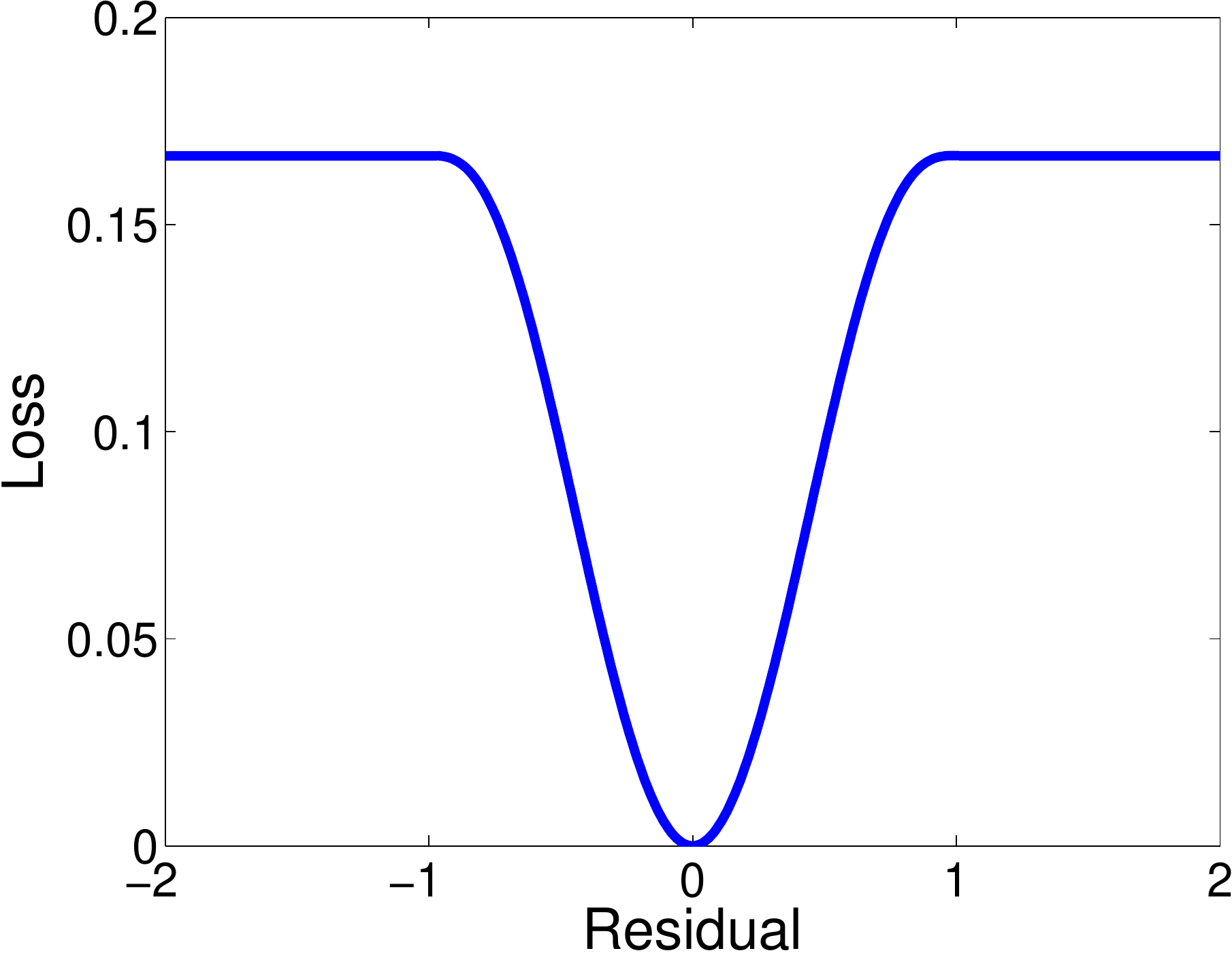}
   \includegraphics[width=0.2\textwidth]{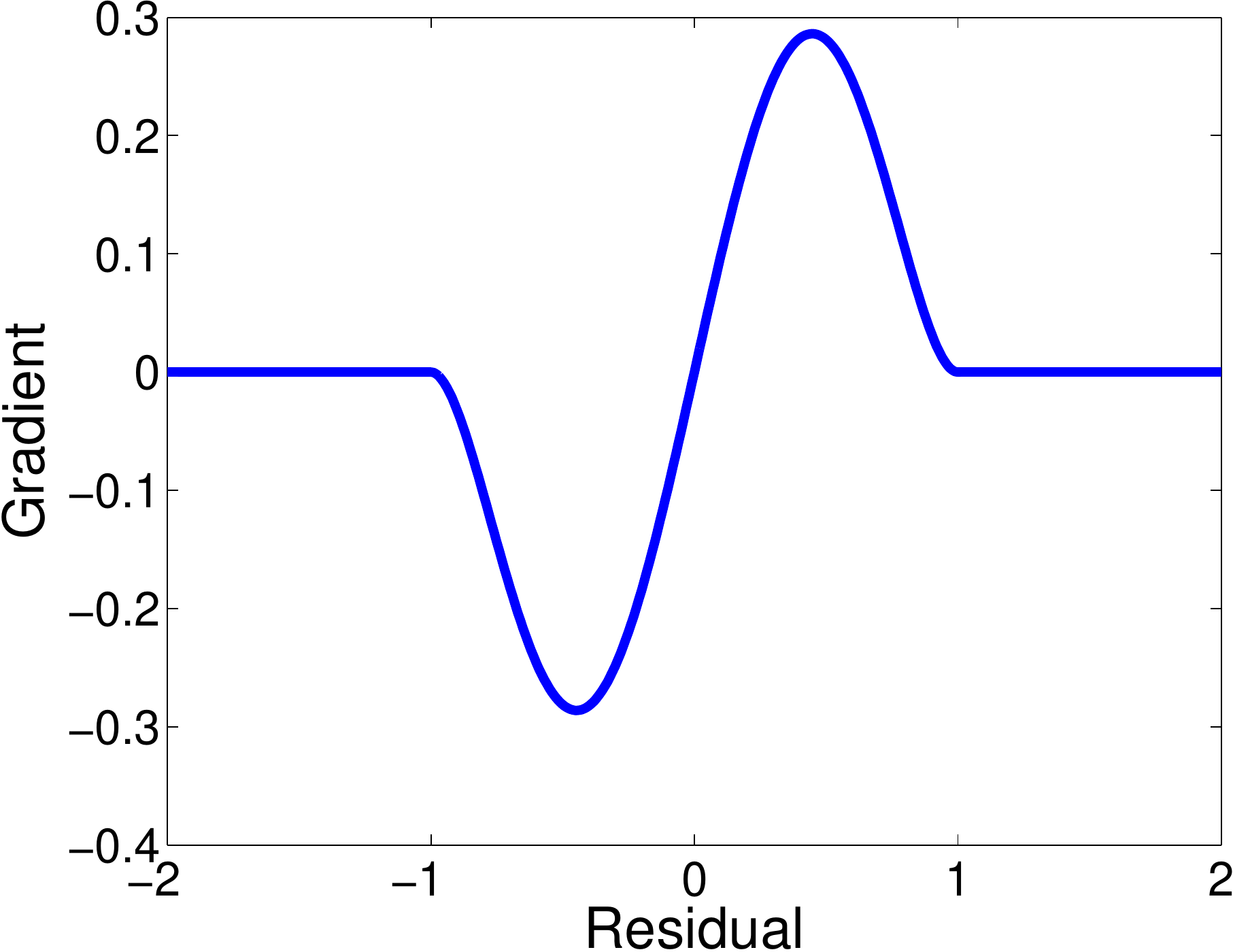}
\caption{An illustration of the Turkey's biweight loss and its first order derivative.}  \label{fig:turkeyLoss}
\end{figure}

To fully exploit the flexibility of the proposed task-specific CRF parameter learning,
which allows us to choose arbitrary learning criterion, here we
  consider a non-convex, robust regression loss function, namely, the Turkey's biweight loss:
\begin{equation}\label{eq:turkeyloss}
\rho(r_p) = \left\{
\begin{array}{l l}
     \frac{c^2}{6} \big[1- (1-\frac{r_p^2}{c^2})^3 \big],&\text{if}\; |r_p| < c \\
    \frac{c^2}{6}, &\text{otherwise.}
\end{array}\right.
\end{equation}
Here $c$ is a pre-defined constant.
Fig.~\ref{fig:turkeyLoss} shows the Turkey's biweight loss and its first-order derivative.

%% file: exp.tex
 \begin{figure} \center
   \scalebox{0.55}
   {
\begin{tabular}{cccc}

      \includegraphics[width=0.2\textwidth, height=0.16\textwidth]{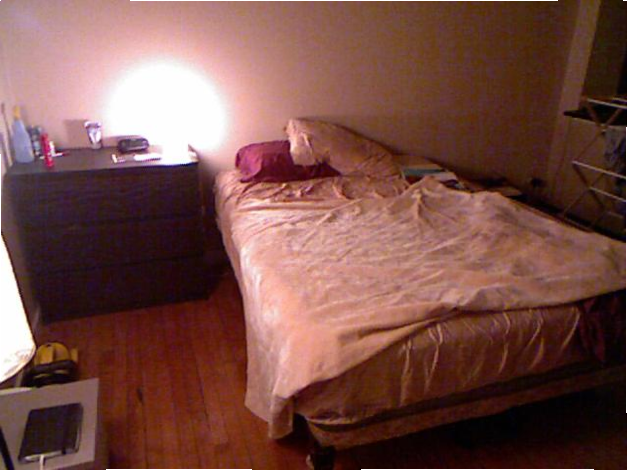}
      &\includegraphics[width=0.2\textwidth, height=0.16\textwidth]{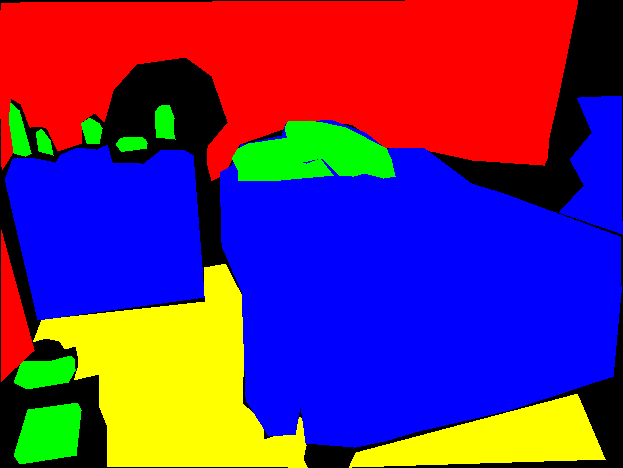}
      &\includegraphics[width=0.2\textwidth, height=0.16\textwidth]{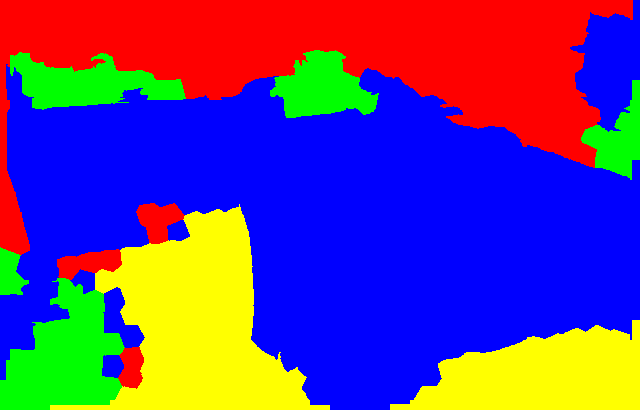}
      &\includegraphics[width=0.2\textwidth, height=0.16\textwidth]{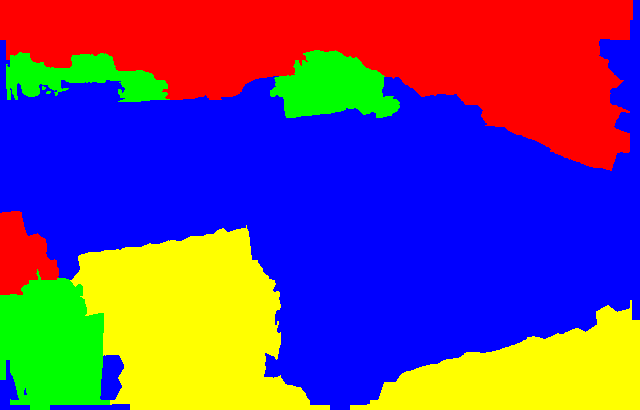}    \\

      \includegraphics[width=0.2\textwidth, height=0.16\textwidth]{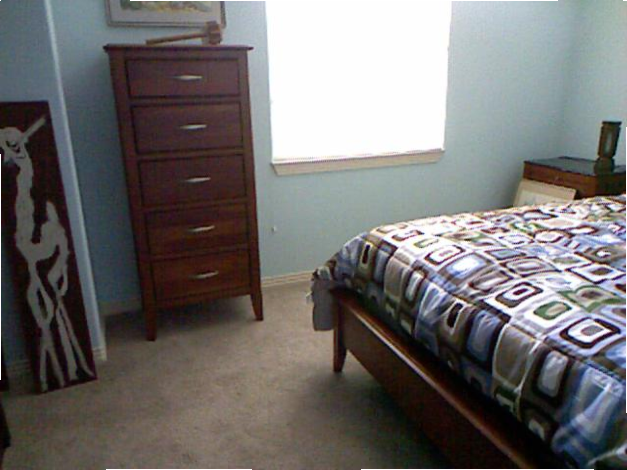}
      &\includegraphics[width=0.2\textwidth, height=0.16\textwidth]{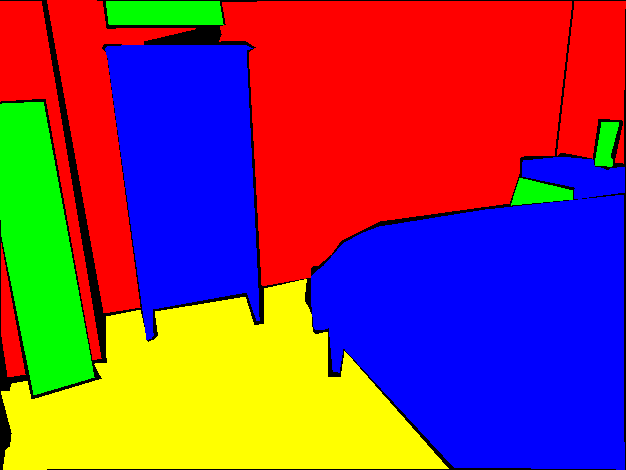}
      &\includegraphics[width=0.2\textwidth, height=0.16\textwidth]{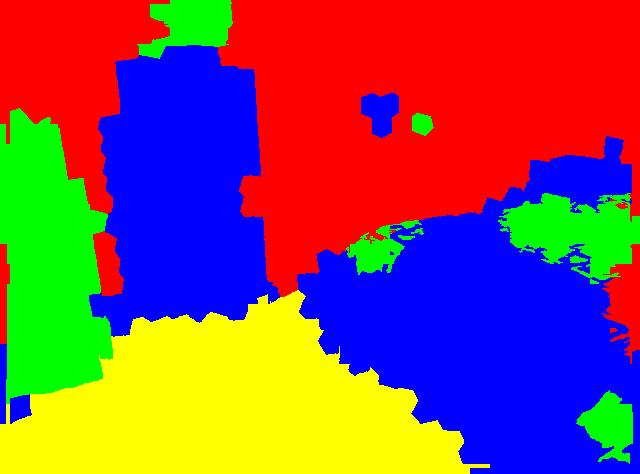}
      &\includegraphics[width=0.2\textwidth, height=0.16\textwidth]{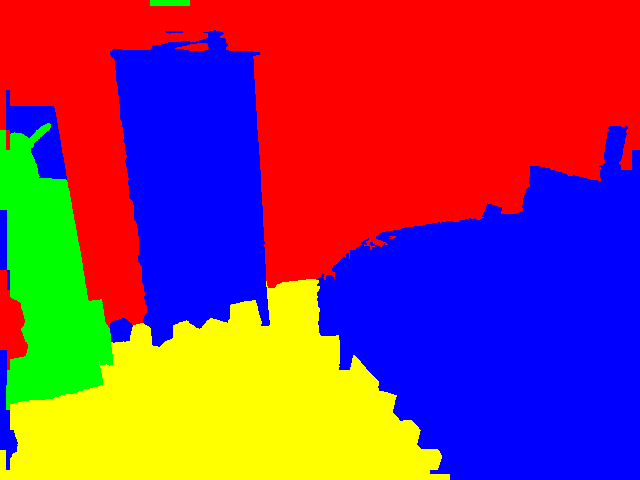}  \\

      \includegraphics[width=0.2\textwidth, height=0.16\textwidth]{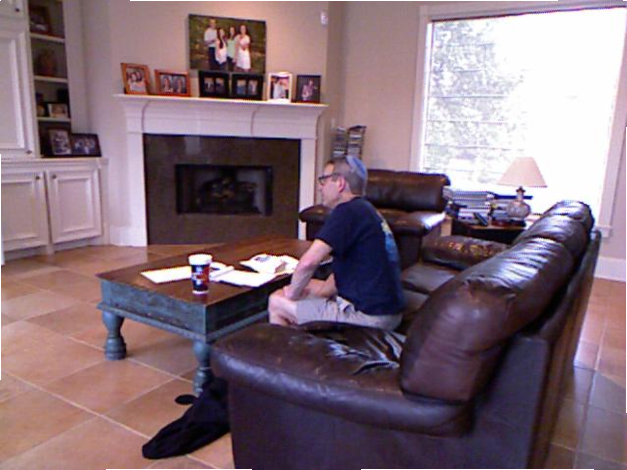}
      &\includegraphics[width=0.2\textwidth, height=0.16\textwidth]{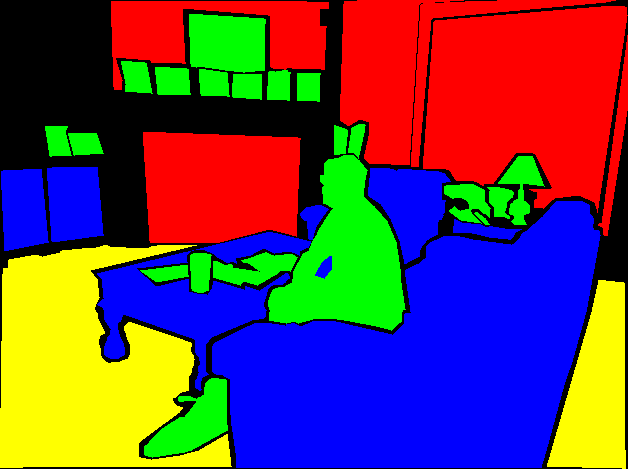}
      &\includegraphics[width=0.2\textwidth, height=0.16\textwidth]{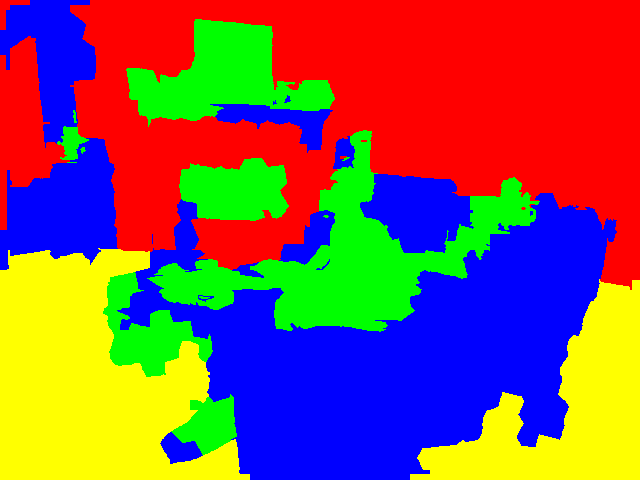}
      &\includegraphics[width=0.2\textwidth, height=0.16\textwidth]{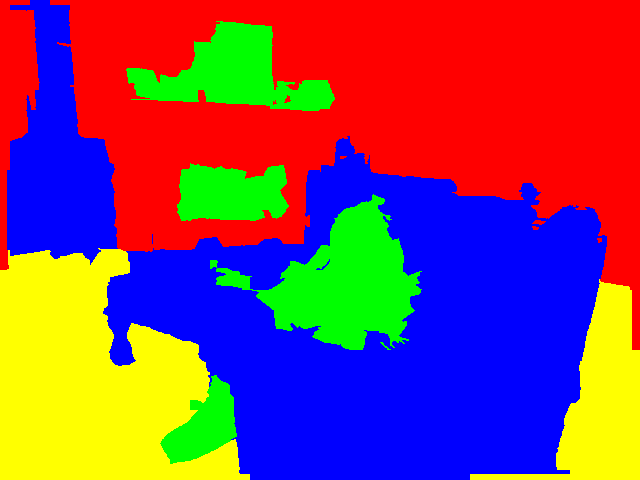}    \\

      \includegraphics[width=0.2\textwidth, height=0.16\textwidth]{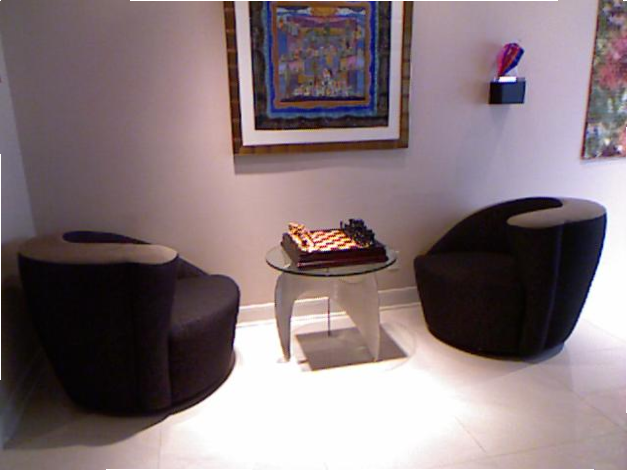}
      &\includegraphics[width=0.2\textwidth, height=0.16\textwidth]{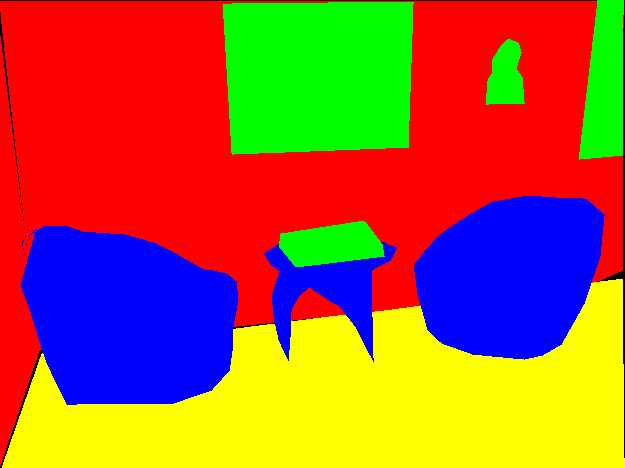}
      &\includegraphics[width=0.2\textwidth, height=0.16\textwidth]{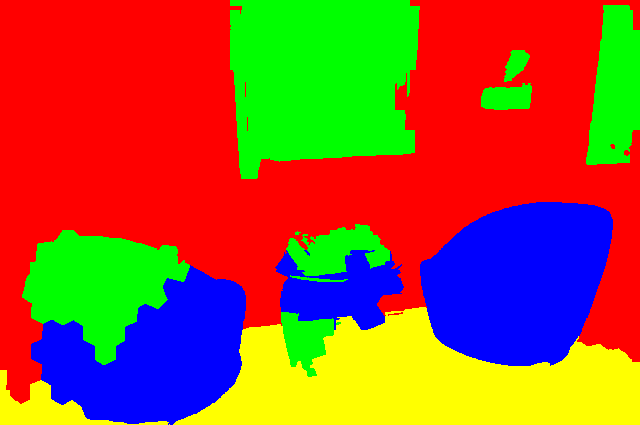}
      &\includegraphics[width=0.2\textwidth, height=0.16\textwidth]{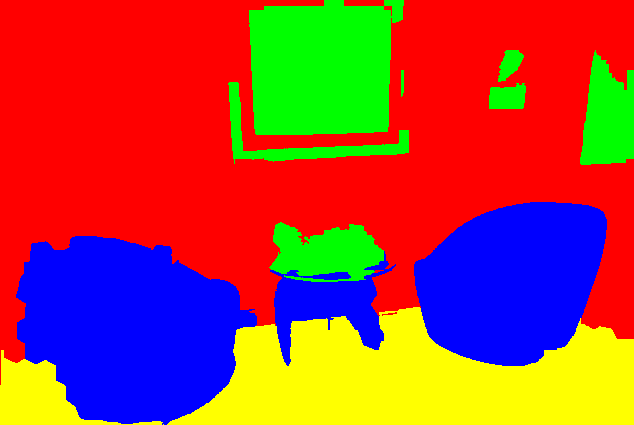}     \\

      \includegraphics[width=0.2\textwidth, height=0.16\textwidth]{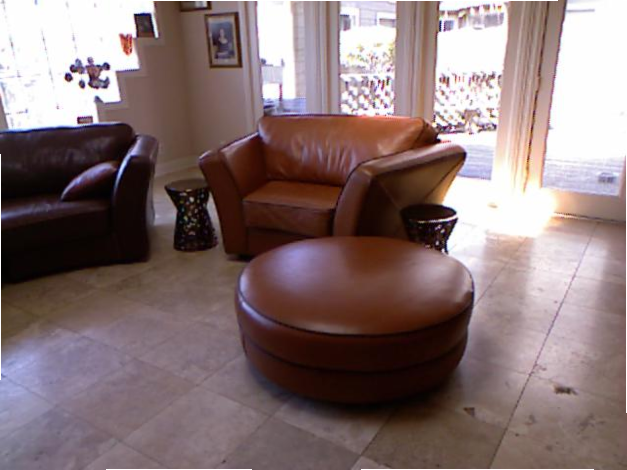}
      &\includegraphics[width=0.2\textwidth, height=0.16\textwidth]{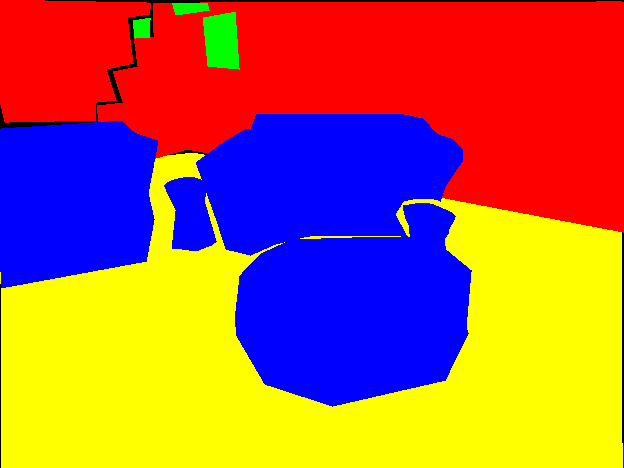}
      &\includegraphics[width=0.2\textwidth, height=0.16\textwidth]{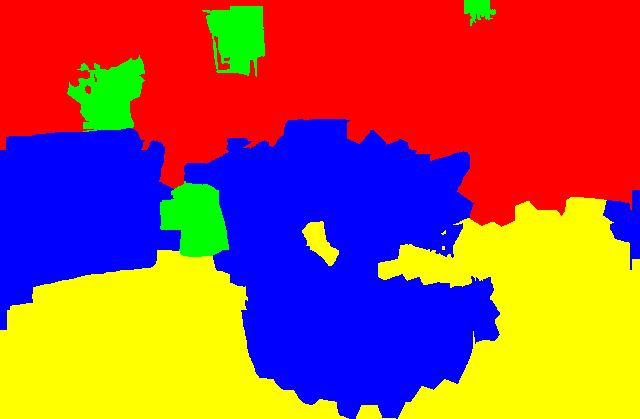}
      &\includegraphics[width=0.2\textwidth, height=0.16\textwidth]{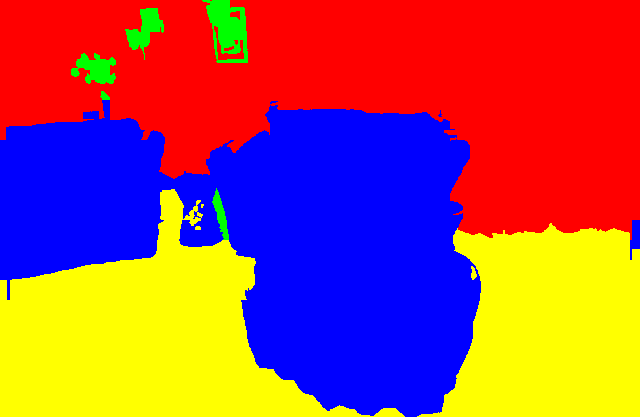}    \\

Test image &Ground-truth &Unary prediction &Our full model \\

\end{tabular}
}
\caption{Semantic labelling (4-class) examples on the NYU v2 dataset. Compared to the unary predictions, our full model yields more
accurate and smoother labelling.}  \label{fig:nyud2_suppl}
\end{figure}

\begin{figure*} [t] \center
{
	\includegraphics[width=0.28\textwidth]{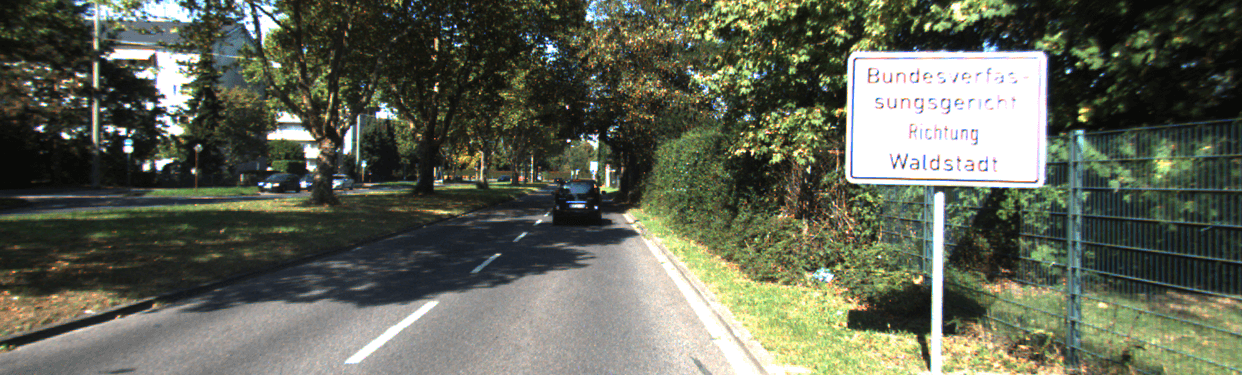}
	\includegraphics[width=0.28\textwidth]{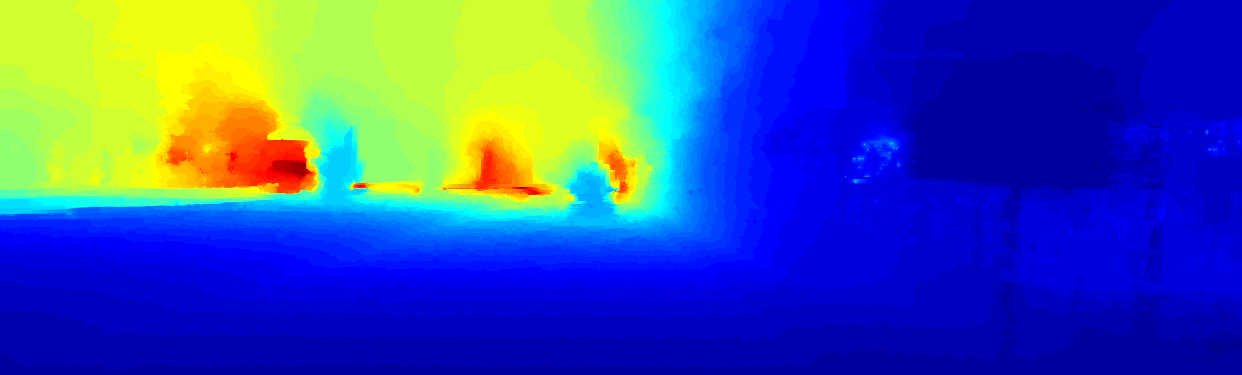}
	\includegraphics[width=0.28\textwidth]{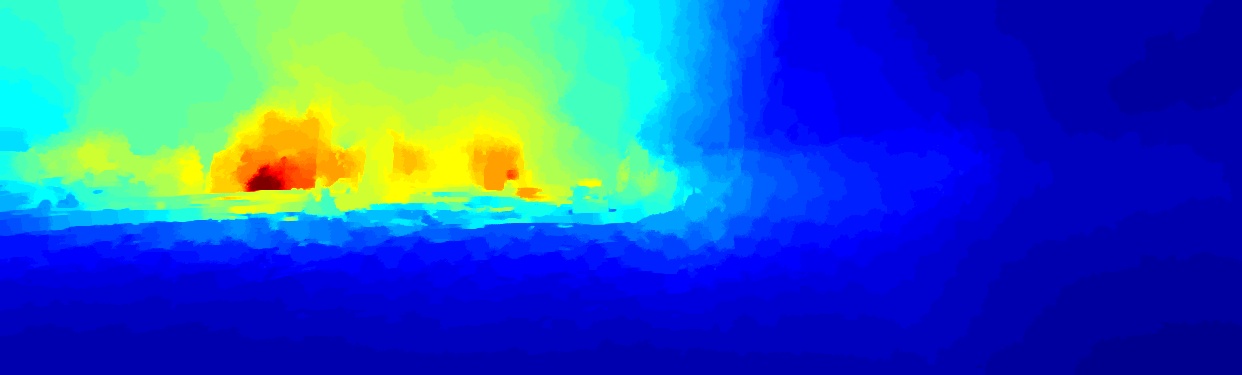}\\

	\includegraphics[width=0.28\textwidth]{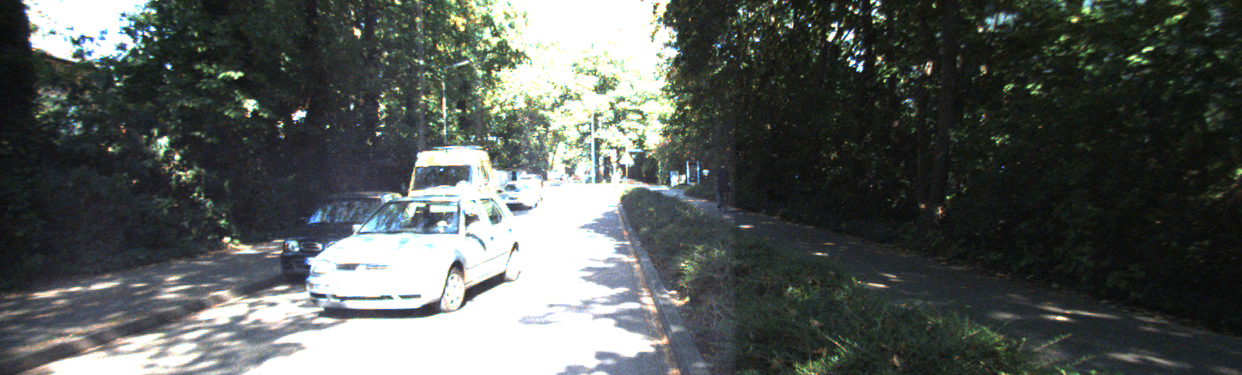}
	\includegraphics[width=0.28\textwidth]{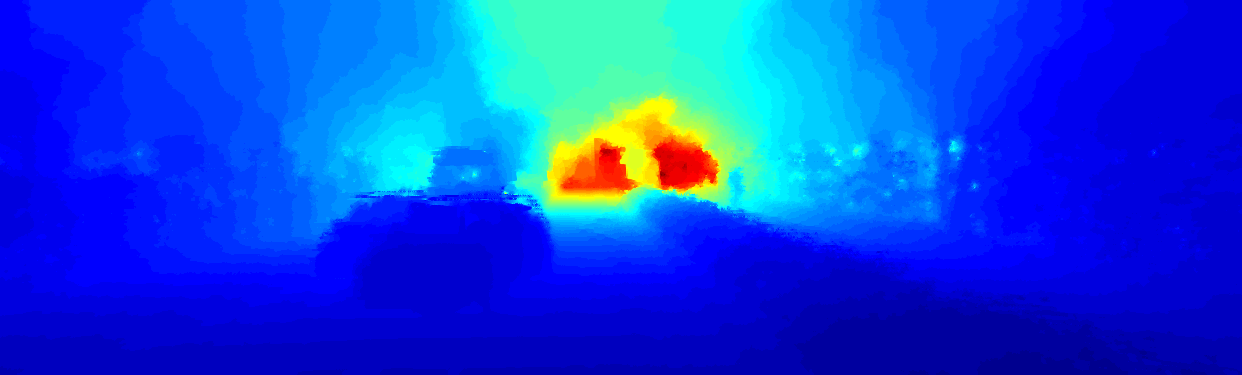}
	\includegraphics[width=0.28\textwidth]{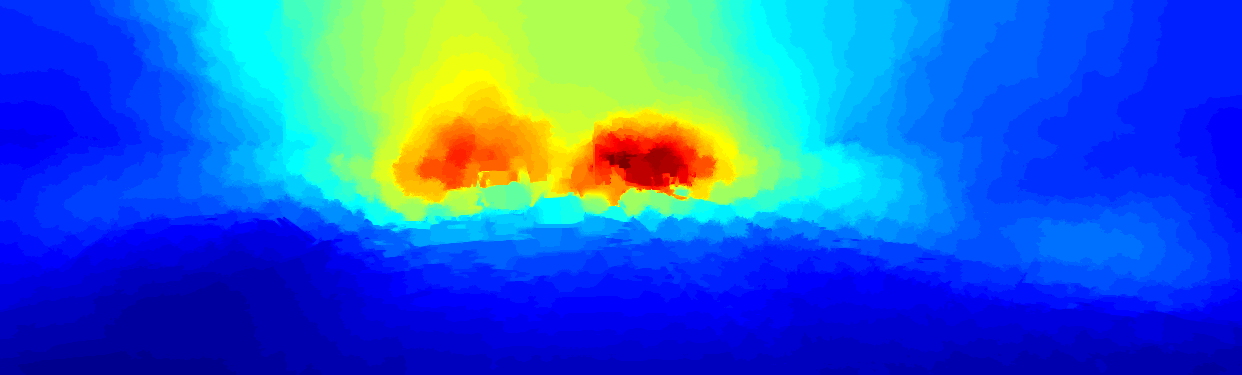}\\

	\includegraphics[width=0.28\textwidth]{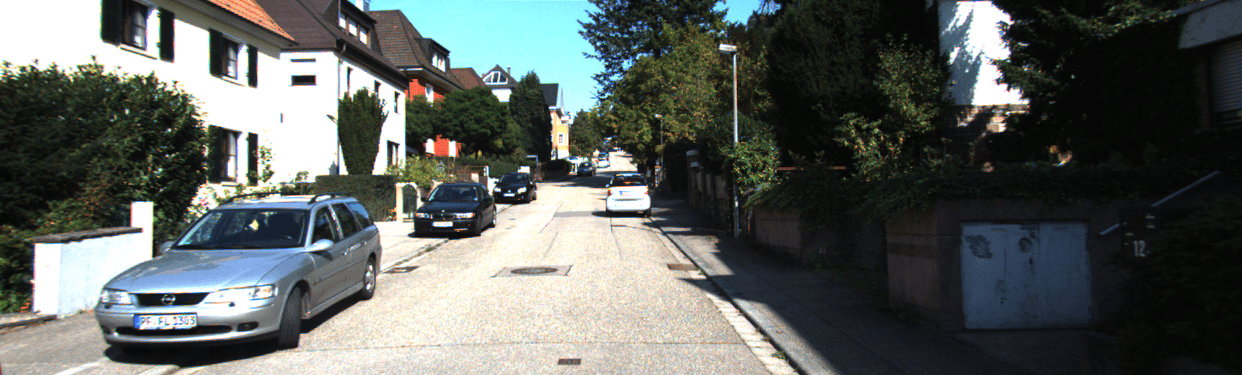}
	\includegraphics[width=0.28\textwidth]{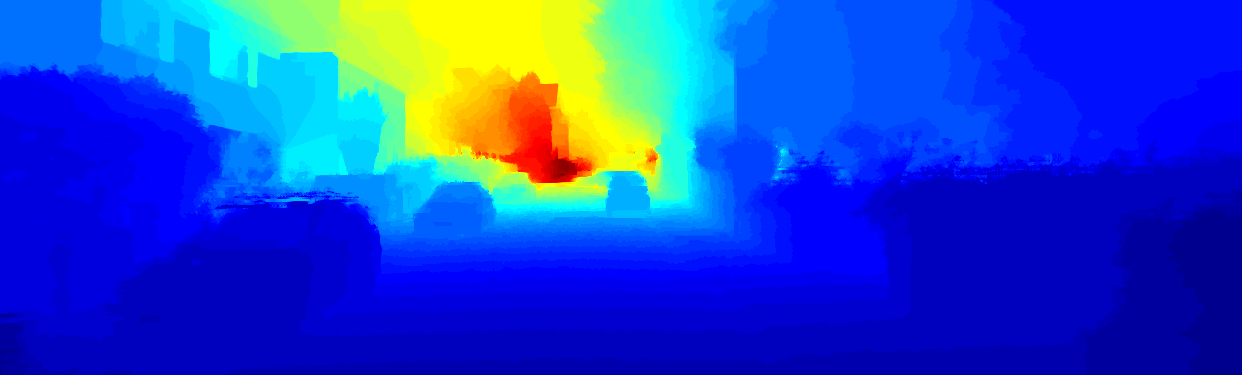}
	\includegraphics[width=0.28\textwidth]{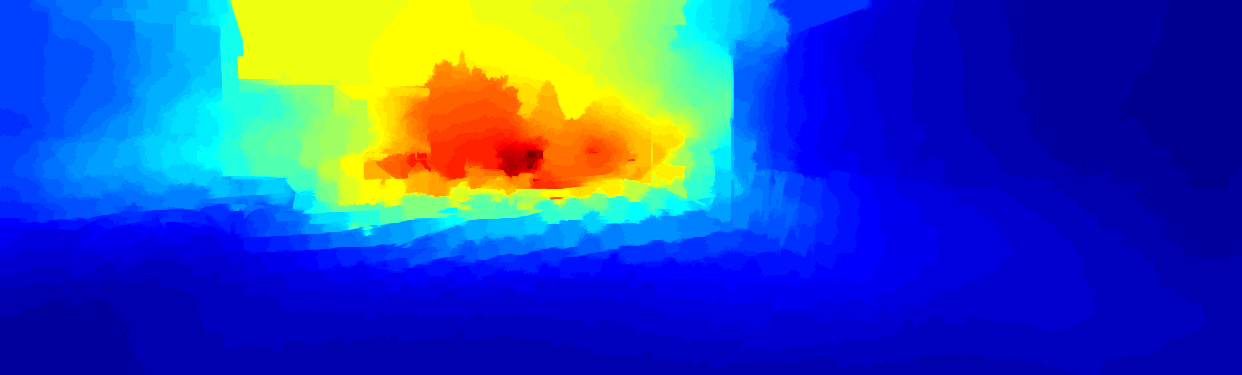}\\

	\includegraphics[width=0.28\textwidth]{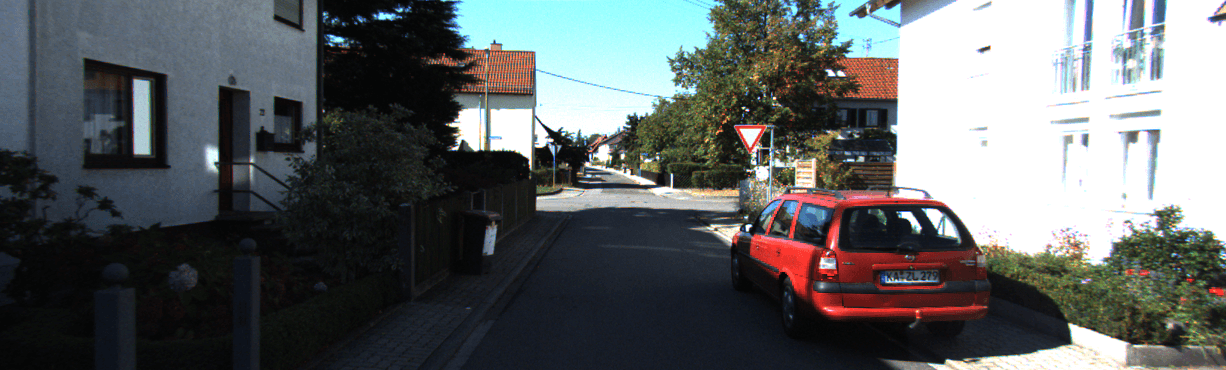}
	\includegraphics[width=0.28\textwidth]{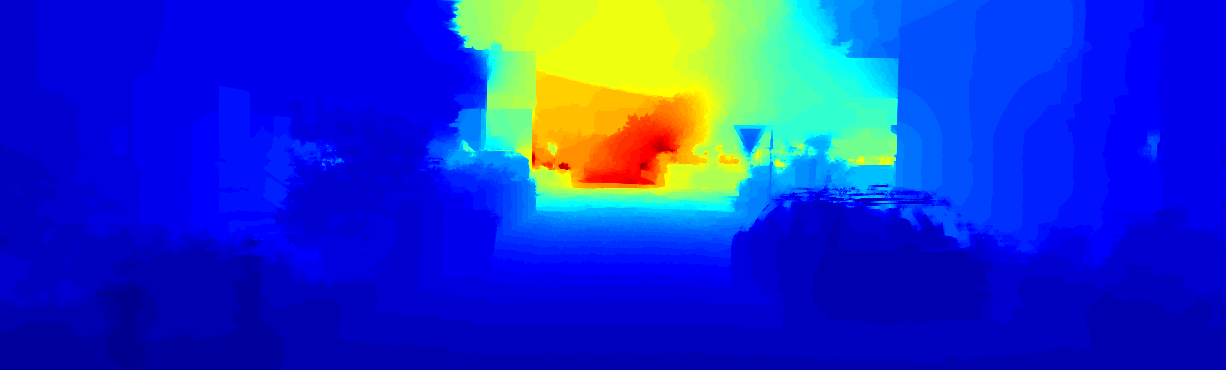}
	\includegraphics[width=0.28\textwidth]{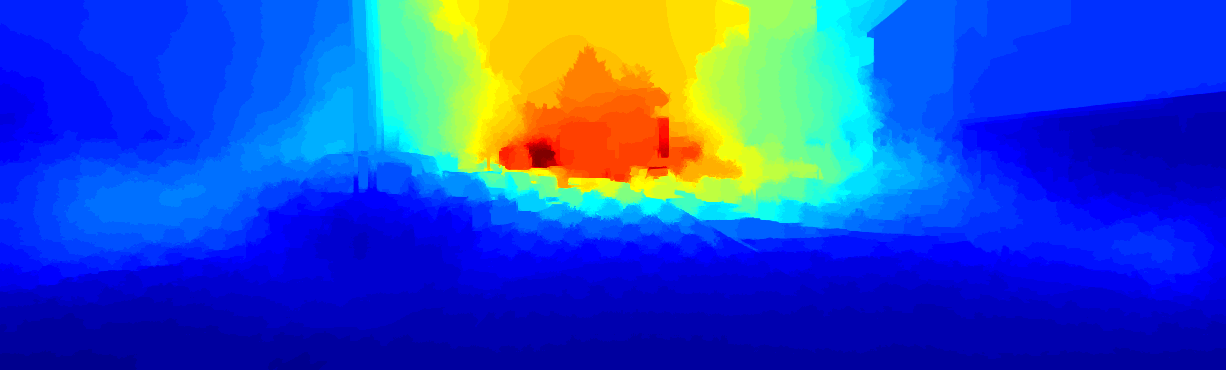}\\

\footnotesize{
Test image \quad\quad\quad\quad\quad\quad\quad\quad\quad\quad\quad\quad
 Ground-truth \quad\quad\quad\quad\quad\quad\quad\quad\quad\quad\quad\quad Our predictions
 }
 \\
}
\caption{Examples of depth predictions on the KITTI dataset (Best viewed on screen).
Depths are shown in log scale and in color (red is far, blue is close).
}  \label{fig:kitti}
\end{figure*}

\begin{figure}[h!]  \center
\begin{tabular}{c c}
{
      \includegraphics[width=0.2175\textwidth, height=0.175\textwidth]{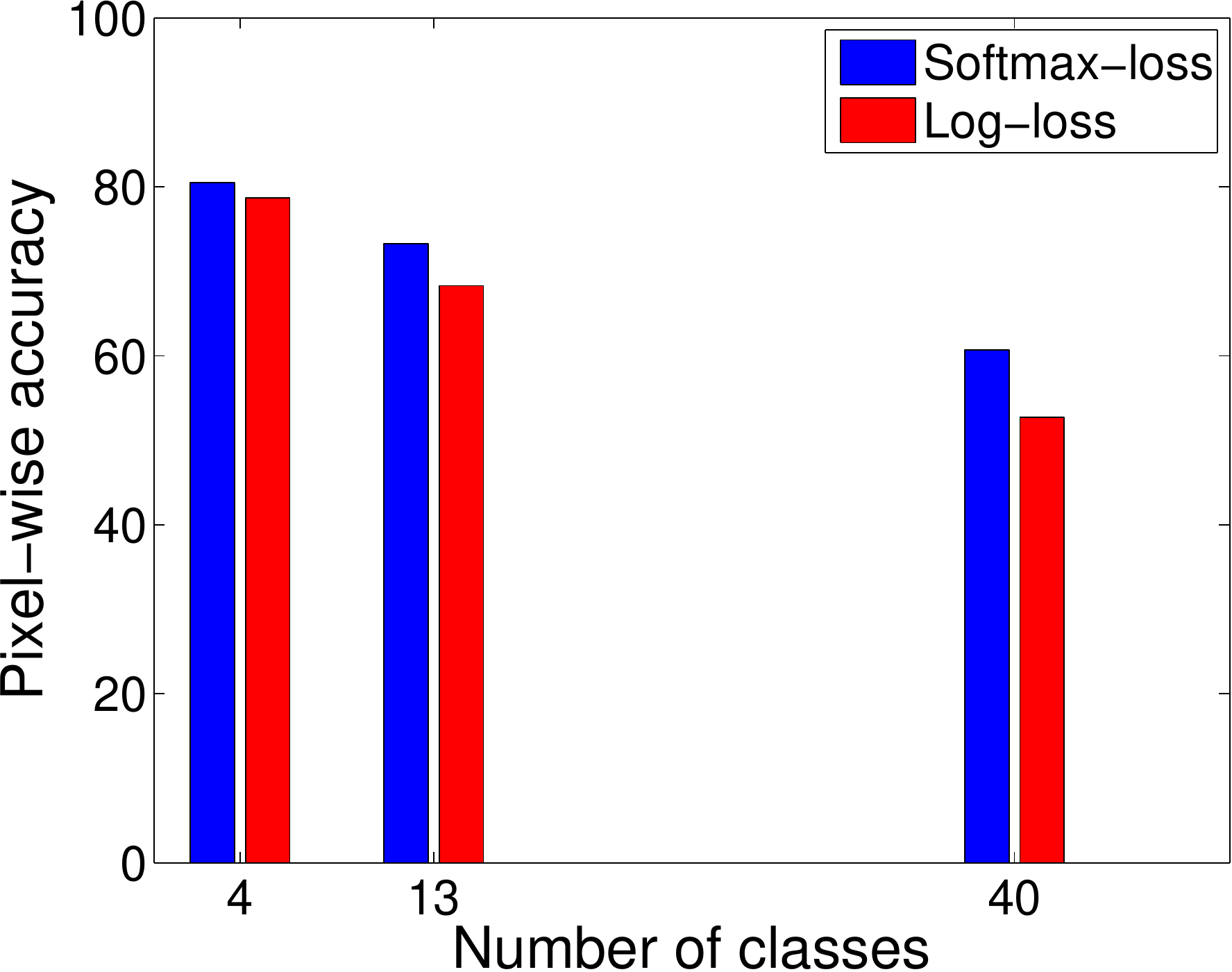}
      \includegraphics[width=0.2175\textwidth, height=0.175\textwidth]{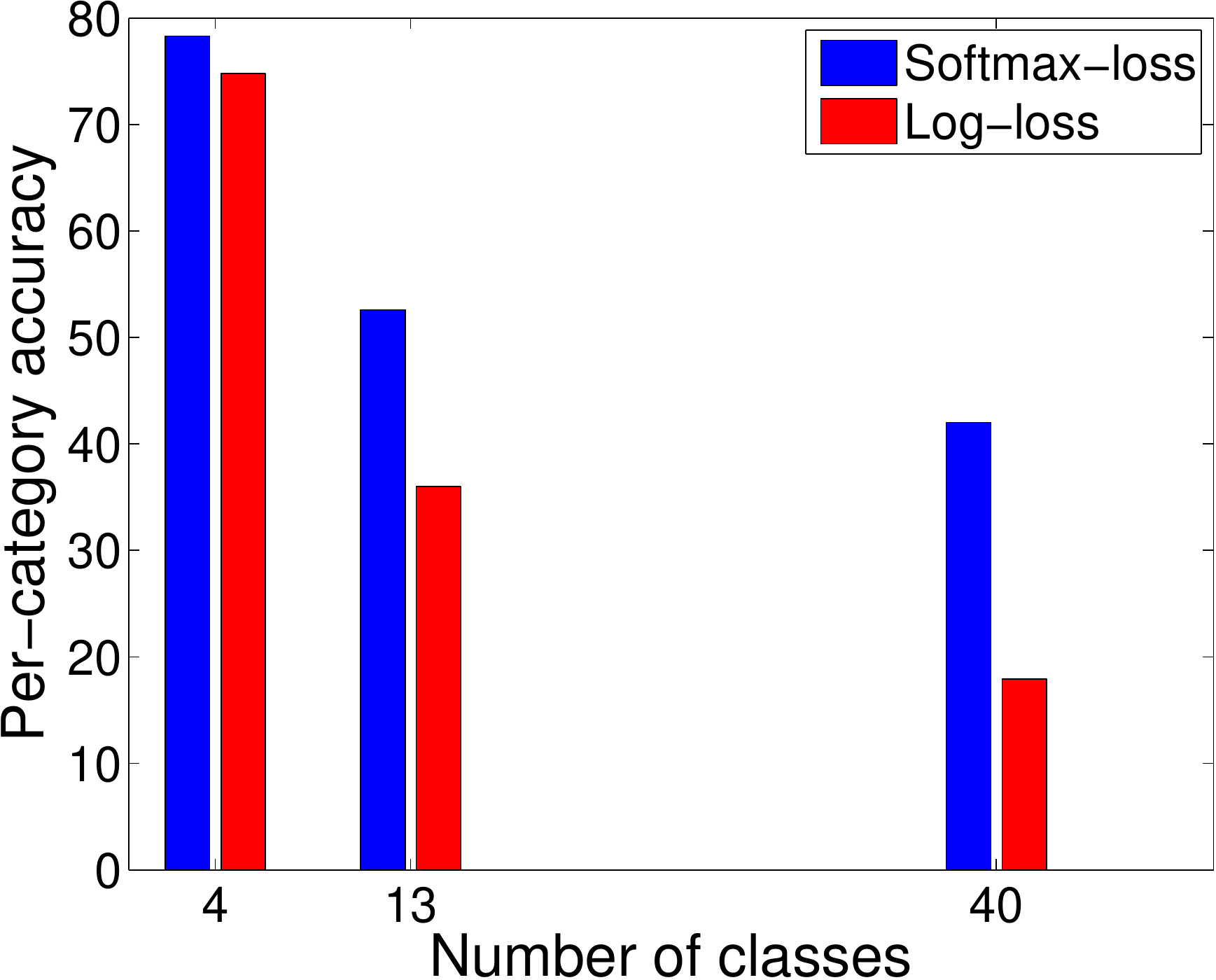}
 }
\end{tabular}
\caption{Performance comparison of the conditional log-likelihood loss and the softmax loss for image segmentation on the NYU v2 dataset.}  \label{fig:acc_class}
\end{figure}

\begin{table*}
\center
\resizebox{.6\linewidth}{!} {
\begin{tabular}{  l |  c  c  c c || c  c  }
\hline
&Ground  &Structure &Furniture &Props  &Pix. Acc.  &Class Acc. \\
\hline
\hline
Gupta \etal \cite{Gupta13} &- &- &- &- &78.0	 &-	 \\
Couprie \etal \cite{Couprie14} (RGB+D)  &87.3 &86.1 &45.3 &35.5 &64.5  &63.5  \\
Khan \etal \cite{Khan15} &87.1 &88.2 &54.7 &32.6 &69.2  &65.6 \\
Muller \etal \cite{Muller14} &\textbf{94.9} &78.9 &71.1 &42.7 &72.3 &71.9 \\
Eigen \etal (RGB) \cite{Eigen15}    &- &- &- &-  &75.3  &- \\
Eigen \etal (RGB+D+N) \cite{Eigen15}	&93.9 &87.9 &79.7 &55.1 &80.6	&79.1	\\
\hline
Ours (unary) &84.3 &84.8 &77.3 &54.6  &77.1  &75.8   \\
Ours (full, RGB) &88.0  &89.8  &77.7  &58.0 &80.8	&78.7	 \\
Ours (full, RGB+D)  &92.9  &\textbf{88.7}  &\textbf{81.7}  &\textbf{61.7}  &\textbf{82.5} &\textbf{81.2} \\
\hline
\end{tabular}
}
\caption{Segmentation results on the NYU v2 dataset (4-class).
  Note that our full model even outperforms  the RGB+D+N model in \cite{Eigen15} which uses additional ground-truth depth and surface normal
  information.
}
\label{tab:seg_nyu4}
\end{table*}

\begin{table} \center
\resizebox{.99\linewidth}{!} {
\begin{tabular}{  l |  c  c  c c}
\hline
&Pix. Acc.  &Class Acc. &Avg. Jaccard  &Freq. Jaccard\\
\hline
\hline
Gupta \etal \cite{Gupta13} &59.1   &28.4   &27.4  &45.6   \\
Gupta \etal \cite{Gupta14} &60.3   &35.1   &28.6  &47.0   \\
Khan \etal \cite{Khan15}  &50.7 &\textbf{43.9}  &- &42.1\\
Long \etal \cite{Long15} (RGB) &60.0 &42.2 &29.2 &43.9 \\
Long \etal \cite{Long15} (RGB+D) &61.5  &42.4 &30.5 &45.5 \\
Eigen \etal \cite{Eigen15} (RGB+D+N)      &62.9     &41.3   &\textbf{30.8}  &47.6   \\
\hline
Ours (unary) &58.8    &38.2   &27.1  &44.6\\
Ours (full, RGB) &62.6    &42.2   &30.5  &\textbf{49.1} \\
Ours (full, RGB+D) &\textbf{63.1}	 &39.0 &	29.5  &48.4 \\
\hline
\end{tabular}
}
\caption{State-of-the-art comparison on the NYU v2 dataset (40-class).}
\label{tab:seg_nyu40}
\end{table}

\begin{table*}
\resizebox{.99\linewidth}{!} {
\begin{tabular}{  l |  c  c  c c c  c  c c c c c  c  c c c  c  c c c c }
\hline
 &wall &floor &cabinet    &bed    &chair    &sofa    &table    &door    &window  &bookshelf
 &picture  &counter &blinds &desk &shelves &curtain &dresser &pillow &mirror &floor mat  \\
\hline
Gupta \etal \cite{Gupta14} &68.0 &81.3 &44.9 &\textbf{65.0} &\textbf{47.9} &\textbf{47.9} &29.9 &20.3 &32.6 &18.1 &40.3 &51.3 &42.0 &\textbf{11.3} &3.5 &29.1 &\textbf{34.8} &\textbf{34.4} &16.4 &\textbf{28.0} \\
Eigen \etal \cite{Eigen15} &\textbf{68.2} &\textbf{83.3}  &45.0  &51.9  &46.0 &40.4 &\textbf{32.0} &17.6 &33.0 &31.5 &45.0 &\textbf{53.4} &43.9 &\textbf{11.3} &\textbf{8.9} &\textbf{35.4} &26.3 &31.0 &\textbf{32.6}  &27.4 \\
Ours &67.7  &75.9  &\textbf{46.8}  &52.2  &43.8  &46.3  &29.0  &\textbf{27.3}  &\textbf{33.2}  &\textbf{35.9}  &\textbf{48.6}  &41.9  &\textbf{46.6}  &9.0  &7.8  &32.9  &28.9  &24.6  &18.1  &23.0 \\
\hline
\hline
\end{tabular}
}
\resizebox{.99\linewidth}{!} {
\begin{tabular}{l | c  c  c c c  c  c c c c c  c  c c c  c  c c c c}
&clothes &ceiling    &books    &fridge    &tv    &paper    &towel  &s-curtain   &box &w-board    &person    &n-stand    &toilet    &sink    &lamp  &bathtub    &bag    &other-struc  &other-fur    &other-prop \\
\hline
Gupta \etal \cite{Gupta14}   &4.7 &60.5 &6.4 &14.5 &31.0 &14.3 &16.3 &4.2 &2.1 &14.2 &0.2 &\textbf{27.2} &\textbf{55.1} &37.5 &\textbf{34.8} &\textbf{38.2} &0.2 &7.1 &6.1 &23.1 \\
Eigen \etal \cite{Eigen15} &12.9 &\textbf{75.2} &16.6 &15.7 &32.3 &\textbf{20.3} &14.8 &21.7 &4.5 &6.6 &27.5 &24.2 &44.8 &\textbf{39.4} &33.6 &29.1 &1.5 &11.7 &6.8 &\textbf{29.4}  \\
Ours  &\textbf{15.6}  &52.6  &\textbf{19.3}  &\textbf{22.3}  &\textbf{40.7}  &17.2  &\textbf{22.5}  &\textbf{24.1}  &\textbf{5.6}  &\textbf{46.9}  &\textbf{47.6}  &19.9  &50.6  &37.8  &24.1  &10.3  &\textbf{4.2}  &\textbf{13.4}  &\textbf{10.4}  &26.3 \\
\hline
\end{tabular}
}
\caption{The per-class pixel-wise Jaccard index for each of the 40 categories on the NYU v2 dataset.
Our method performs the best on 18 out of the 40 classes. Note that Eigen \etal \cite{Eigen15}
use extra surface normal information, and their network takes multi-scale input images.
}
\label{tab:seg_nyu40_cat}
\end{table*}

\begin{figure}  \center
\begin{tabular}{c c}
{
      \includegraphics[width=0.2175\textwidth, height=0.175\textwidth]{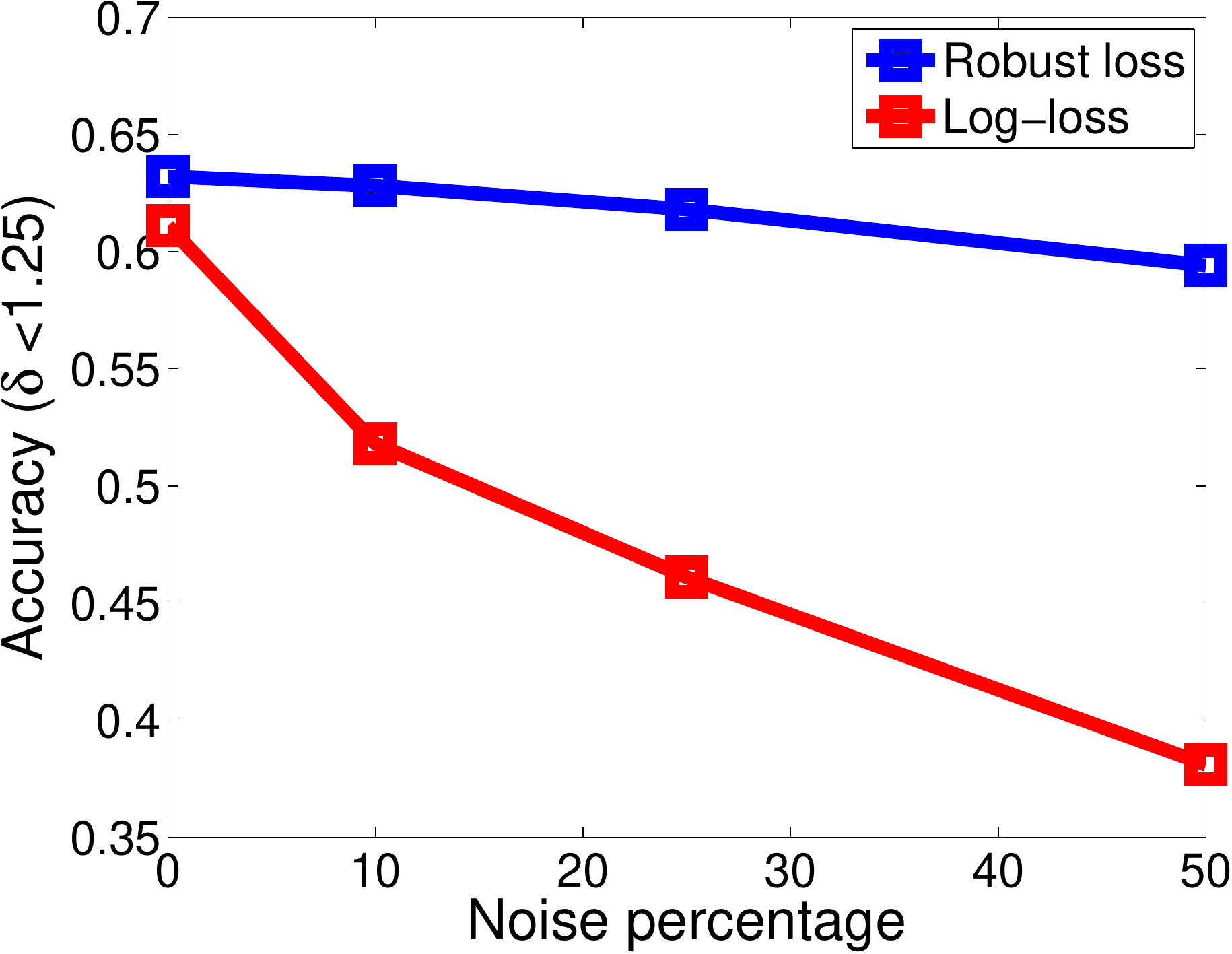}
      \includegraphics[width=0.2175\textwidth, height=0.175\textwidth]{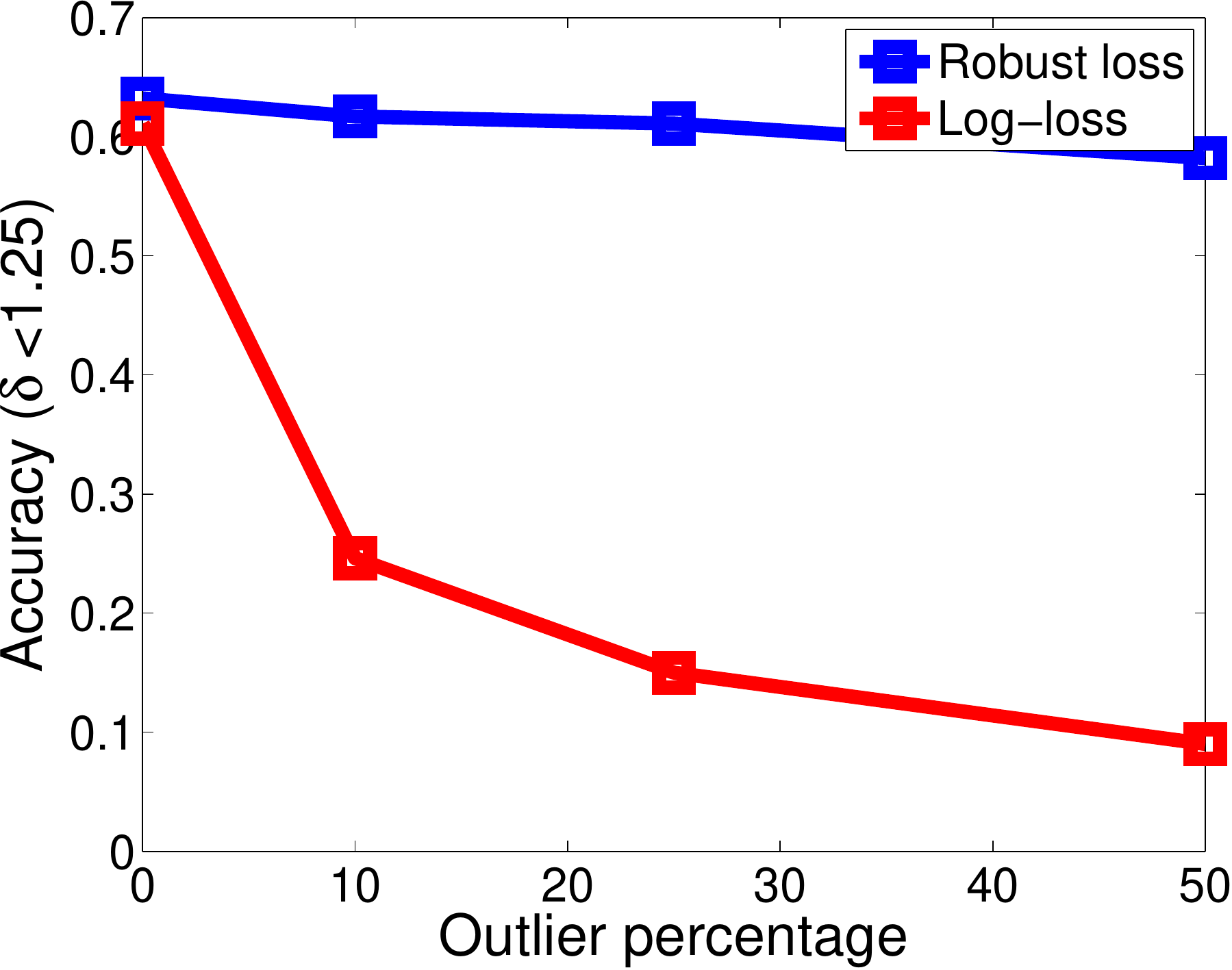}
 }
\end{tabular}
\caption{Prediction accuracies of the log-likelihood loss and the robust regression loss for depth estimation with respect to different percentages of noises/outliers on the NYU v2 dataset.}  \label{fig:acc_noise}
\end{figure}

\begin{table*}
\center
\resizebox{.55\linewidth}{!} {
\begin{tabular}{|  l | l |  c  c  c | c  c  c |}
\hline
\multirow{3}{*}{{{Noise level}}} &\multirow{3}{*}{{{Method}}} &\multicolumn{3}{c|}{Error} &\multicolumn{3}{c|}{Accuracy} \\
& &\multicolumn{3}{c|}{(lower is better)} &\multicolumn{3}{c|}{(higher is better)} \\
\cline{3-8}
& &rel &log10 &rms &$\delta < 1.25$ &$\delta < 1.25^2$ &$\delta < 1.25^3$  \\
\hline
\hline
\multirow{2}{*}{{{$0\%$}}} &Log loss  &0.223 &0.094 &0.835  &0.611& 0.891 &0.973 \\
 &Robust loss &\textbf{0.219} &\textbf{0.091} &\textbf{0.787}  &\textbf{0.632} &\textbf{0.894}  &\textbf{0.974}	\\
\hline
\multirow{2}{*}{{{$10\%$ noise}}}  &Log loss	&0.263  &0.115 &1.023 &0.518 &0.815 &0.945  \\
&Robust loss  &\textbf{0.222}  &\textbf{0.092} &\textbf{0.790} &\textbf{0.628} &\textbf{0.896} &\textbf{0.975}  \\
\hline
\multirow{2}{*}{{{$25\%$ noise}}}  &Log loss 	 &0.369 &0.130  &1.256  &0.461  &0.764 &0.915\\
&Robust loss  &\textbf{0.218}  &\textbf{0.093} &\textbf{0.815} &\textbf{0.618} &\textbf{0.894} &\textbf{0.976} \\
\hline
\multirow{2}{*}{{{$10\%$ outlier}}}  &Log loss	&0.786 &0.211 &2.152  &0.246 &0.477 &0.685 \\
&Robust loss &\textbf{0.221} &\textbf{0.092}	&\textbf{0.812}	&\textbf{0.617}  &\textbf{0.895}  &\textbf{0.974} \\
\hline
\multirow{2}{*}{{{$25\%$ outlier}}}  &Log loss &0.951 &0.261 &2.503 &0.150 &0.371 &0.617 \\
&Robust loss &\textbf{0.235} &0\textbf{0.094} &\textbf{0.819} &\textbf{0.611} &\textbf{0.888} &\textbf{0.972} \\
\hline
\end{tabular}
}
\caption{Comparisons of depth estimation performance of different loss functions with respect to different noise/outlier ratios on the NYU v2 dataset.} \label{tab:depth_noisy}
\end{table*}

\begin{table}
\center
\resizebox{0.99\linewidth}{!} {
\begin{tabular}{ | l |  c  c  c | c  c  c |}
\hline
\multirow{3}{*}{{{Method}}} &\multicolumn{3}{c|}{Error} &\multicolumn{3}{c|}{Accuracy} \\
&\multicolumn{3}{c|}{(lower is better)} &\multicolumn{3}{c|}{(higher is better)} \\
\cline{2-7}
&rel &log10 &rms &$\delta < 1.25$ &$\delta < 1.25^2$ &$\delta < 1.25^3$  \\
\hline
\hline
Saxena \etal \cite{make3d_pami09} &0.280  &-  &8.734  &0.601 &0.820  &0.926 \\
Eigen \etal \cite{dcnn_nips14} &\textbf{0.190} &- &7.156 &\textbf{0.692} &\textbf{0.899} &\textbf{0.967} \\
Liu \etal \cite{pami15} &0.217 &0.092 &7.046  &0.656 &0.881 &0.958  \\
\hline
Ours &0.203	&\textbf{0.086}	&\textbf{6.427} &0.684  &0.894  &0.965	 \\
\hline
\end{tabular}
}
\caption{Comparisons of depth estimation performance on the KITTI dataset.
The results of \cite{dcnn_nips14} are obtained by using millions of training images while ours are obtained by training on 700 images.}
\label{tab:depth_kitti}
\end{table}

\section{Experiments}

To demonstrate the effectiveness of the proposed method,
we perform experiments on the tasks of multi-class semantic labelling and robust depth estimation from single images.
The first one is discrete-valued labelling and the second one is continuously-valued labelling.

\paragraph{Implementation details}
We implement our method based on the popular \cnn toolbox  MatConvNet\footnote{http://www.vlfeat.org/matconvnet/}.
The network training is performed on a standard desktop with an NVIDIA GTX 780 GPU.
We set the momentum as $0.9$ and the weight decay parameter as $\lambda=0.0005$.
The first $5$ convolution blocks and the first layer of the $6$th convolution block of the unary network in Fig. \ref{fig:cnn_net} are initialized from the VGG-16 \cite{vgg16} model.
All layers are trained using back-propagation.
The learning rate for randomly initialized layers is set to $10^{-5}$; for
VGG-16 initialized layers it is set to a smaller value $10^{-6}$.
The hyperparameter of the position Gaussian kernel is set to $\gamma=0.1$.
The parameter $c$ in the Turkey's biweight loss (Eq.~\eqref{eq:turkeyloss}) is set to $1$.
We use SLIC \cite{slic} to generate  $\sim 700$ superpixels for each image.

\subsection{Multi-class semantic labelling}
The semantic labelling experiments are evaluated on the NYU v2 \cite{Silberman12} and the MSRC-21 \cite{Shotton06} datasets.
The NYU v2 dataset consists of 795 images for training and 654 images for test
(we use the standard training/test split provided with the dataset).
For the semantic label ground-truth, we use the 4 and 40 classes described in \cite{Silberman12}, \cite{Gupta13} respectively.
The performance are evaluated using the commonly applied metrics as in \cite{Gupta14,Couprie14,Khan15,Eigen15},
namely, pixel-wise accuracy (Pix.\ Acc.) and average per-class accuracy (Class Acc.).
For the 40-class segmentation task, we also report the average Jaccard index (Avg.\ Jaccard)
and the average pixel-frequency weighted Jaccard index (Freq.\ Jaccard) for each class.
The MSRC-21 dataset is a popular multi-class segmentation benchmark with 591 images containing objects from 21 categories.
We follow the standard split to divide the dataset into training/validation/test subsets.

{\bf NYU v2}
Since ground-truth depth maps are provided along with this dataset, we exploit depth cues as an additional setting (denoted as RGB+D).
Specifically, we replicate the depth channel into 3 duplicates and regard them as images to input into our framework.
The \cnn features from the RGB channels and the depth channels are concatenated after the superpixel pooling layer.
The compared results together with the per-class accuracy on the NYU v2 4-class segmentation task are reported in Table \ref{tab:seg_nyu4}.
As we can see, {\em our full model with incorporated depth cues outperforms state-of-the-art methods by a considerable margin}.
It is worth mentioning that our method outperforms Eigen \etal \cite{Eigen15}, which used RGB images together with depths and surface normals.
Compared with the unary-only model, our full model which jointly learns the unary and the pairwise networks shows consistent
improvements in terms of all the metrics, demonstrating that the continuous \crf model indeed improves discrete labelling.
Fig. \ref{fig:nyud2_suppl} shows some qualitative examples of our method together with the unary baseline.
Compare to the unary predictions, our method yields more accurate and smoother labellings.

Table \ref{tab:seg_nyu40} summarizes the results on the 40-class segmentation task.
Our method generally achieves better or comparable performance against state-of-the-art methods \cite{Long15,Eigen15}.
Specifically, with RGB only as the input, our method outperforms \cite{Long15} using similar settings.
Note that in \cite{Long15}, Long \etal achieved better results by using HHA encoding,  which reveals that similar strategies can be applied here to further improve our results.
In \cite{Eigen15}, Eigen \etal have used RGB images together with the ground-truth depths and surface normals as the input to their network training.
Yet, our method achieves comparable performance.
With depth channels incorporated, our method obtains the best global labelling accuracy.
Our full model again shows improvements over the unary only model, demonstrating the usefulness of the joint learning scheme.
The per-class pixel-wise Jaccard indexes of the 40 object categories are presented in Table \ref{tab:seg_nyu40_cat},
with our method achieves the best results on 18 out of the 40 classes.

{\bf MSRC-21}
Table \ref{tab:seg_msrc} reports the segmentation results on the MSRC-21 dataset.
As we can see, our results performs comparable to state-of-the-art methods \cite{Roy14}, \cite{pr15}.

\begin{table}[h!] \center
\resizebox{.75\linewidth}{!} {
\begin{tabular}{  l |  c  c  }
\hline
&Pix.\ Acc.\  &Class Acc.\   \\
\hline
\hline
Lucchi \etal \cite{Lucchi13}  &83.7  &78.9 \\
Roy \etal \cite{Roy14} &91.5$^*$  &- \\
Liu \etal \cite{pr15}  &88.5 &86.7 \\
Liu \etal  \cite{pr15} (co-occur)  &91.1$^*$ &\textbf{90.5}$^*$ \\
\hline
Ours (softmax loss, unary) &89.6	 &86.4  \\
Ours (softmax loss, full) &\textbf{91.4}    &89.0  \\
\hline
\end{tabular}
}
\caption{Segmentation results on the MSRC-21 dataset.
  Note that the results marked with  $ ^*$    are achieved by using additional
   mutex/co-occurrence information that models repelling   relations.
}
\label{tab:seg_msrc}
\end{table}

\paragraph{Ablation Study}
To show the benefits of the multi-class classification loss over the tradition log-likelihood \crf training,
we perform ablation study on the performance comparisons of the two loss functions with respect to different numbers of classes.
In addition to the 4-class and the 40-class segmentation tasks, we further perform experiments on a 13-class segmentation task by mapping the 40 categories into 12 main classes and a `background' class. Here the network structures are identical and the only difference is the loss function.
Fig.\ \ref{fig:acc_class} shows the comparative performance.
We report both pixel-wise accuracy and per-category accuracy.
    As illustrated in the figure,
    {\em the multi-class soft-max loss considerably outperforms the log-likelihood loss}.
    The discrepancy becomes more significant when the number of classes increases.
    The previously-reported inferior performance of continuous \crf on discrete labelling may be attributed to the use of the 
    traditional maximum likelihood parameter learning.

\subsection{Robust depth estimation}
We perform robust depth estimation on two datasets: the indoor NYU v2 \cite{Silberman12} and the outdoor KITTI \cite{kitti} datasets.
Several commonly used evaluation metrics, \ie, 
relative error (rel), log10 error (log10), root mean square error (rms) and accuracy with different thresholds $\delta$ (details are referred to \cite{cvpr15,pami15}),
are used here to report the results.

\paragraph{NYU v2}
The NYU v2 dataset \cite{nyud2_eccv12} consists of 1449 RGB-D indoor image pairs, with 795 are used for training and 654 are for testing.
We have used the standard training/test split provided along with the dataset.

In addition to performance comparisons on the original dataset, we also add additive white Gaussian noises or outliers to the ground-truth depths.
For the noise case, we add additive white Gaussian noise with $\sigma=0.1$ to the $[0,1]$ normalized ground-truth depths.   A certain percentage of pixels (\eg, 10\%, and 25\% as shown in Table \ref{tab:depth_noisy})
are uniformly sampled to be added the noise.
Likewise, for the outliers, we simply add a large value to the ground-truth of
a certain percentage of uniformly sampled pixels.
Table \ref{tab:depth_noisy} reports the compared results of the log-likelihood loss (denoted as `log loss')
and the robust regression loss with respect to different noise/outlier levels.
As we can observe, the robust regression loss is insensitive to both noises and outliers.
In contrast, the performance of the log-likelihood loss degrades severely when the noise/outlier level increases.
Fig.\ \ref{fig:acc_noise} illustrates the compared prediction accuracies of the two loss functions.
The horizontal axis shows different noise (left plot) or outlier (right plot) percentages,
and the vertical axis shows the depth prediction accuracy with $\delta<1.25$.

\paragraph{KITTI}
The KITTI dataset \cite{kitti} consists of outdoor videos taken from a driving vehicle equipped with a LiDaR sensor.
The ground-truth depths of the KITTI dataset are scattered at irregularly spaced points, which only consists of $\sim 5\%$ pixels of each image, with all others unlabelled.
We here treat those unlabelled regions as outliers and train the network using robust regression loss.
The results are reported by using the same test set, \ie, 697 images from 28 scenes, as provided by Eigen \etal \cite{dcnn_nips14}.
As for the training set, we use the same 700 images that Eigen \etal \cite{dcnn_nips14} have used to train the method of \cite{make3d_pami09}.
The compared results are reported in Table \ref{tab:depth_kitti}.
It can be seen that our method outperforms the most recent work of \cite{pami15} which have used the same training set,
and very similar \cnn structure as ours.
Compared to \cite{dcnn_nips14}, which have used millions of training images, our results are very competitive with the lowest rms error achieved.
Fig.~\ref{fig:kitti} shows some prediction examples of our method.
On both NYU v2 and KITTI datasets, our method using robust loss outperforms  the log-likelihood loss,
even with no noises   deliberately   added, because (a) original data may contain some noises more or less;
(b) more importantly our method directly optimizes the task-specific empirical risk of interest.